\documentclass[twocolumn,secnumarabic,amssymb,nobibnotes,aps,pra]{revtex4-2}
\usepackage{graphicx}
\usepackage{amsmath}  

\usepackage{hyperref} 
\hypersetup{colorlinks=true, linkcolor=black, citecolor=black, urlcolor=blue}

\begin{document}

\title{
	Real-time probabilistic tsunami forecasting via generative AI
}

\author{Yusuke~Oishi}
\affiliation{Fujitsu Research, Fujitsu Limited, Kawasaki, Japan.}

\author{Takashi~Furumura}
\affiliation{Earthquake Research Institute, The University of Tokyo, Tokyo, Japan.}

\author{Fumihiko~Imamura}
\affiliation{International Research Institute of Disaster Science, Tohoku University, Sendai, Japan.}

\begin{abstract} 
Explicit onshore tsunami inundation forecasting can improve public risk awareness, but deterministically predicted inundation boundaries under highly uncertain conditions, such as near-field tsunamis generated by megathrust earthquakes, may falsely imply safety outside the boundaries. Consequently, current warnings primarily target coastal tsunami height, not onshore inundation. 
Although machine learning enables instant inundation predictions, they remain deterministic, lacking uncertainty quantification. Here, we develop a probabilistic ensemble model based on a conditional diffusion model---a type of generative AI---that reconciles accuracy with calibration. 
Validated with the 2011 Tohoku-oki earthquake data, our model faithfully tracks the postearthquake uncertainty decreasing over time while accurately predicting inundation depth and extent. 
Our framework shows that generative AI can shift tsunami forecasting from determinism to probabilism, providing a foundation for next-generation early warning.
\end{abstract}

\maketitle

\section{Introduction}

Tsunamis triggered by megathrust earthquakes are among the most devastating natural threats to coastal communities. Recent sea-level rise and population concentration in coastal areas have further amplified their potential risks \cite{neumann2015future, synolakis2006tsunami}. While tsunami forecasting is fundamental to damage mitigation, an essential challenge remains in adequately handling the inherent uncertainties.

Tsunami forecasting fundamentally relies on incomplete observation data---such as indirect tsunami source estimations from seismic records, estimations of offshore crustal deformation from onshore observations, or limited offshore waveform data---and is thus inherently characterized by uncertainty \cite{angove2019ocean, goda2016uncertainty, selva2021probabilistic, grezio2017probabilistic}. In particular, the nonlinear behavior of tsunamis in shallow waters or  
onshore environments increases prediction uncertainty \cite{titov1997implementation}. Furthermore, for near-field tsunamis, severe time constraints necessitate making predictions with insufficient observation data, rendering the associated uncertainties even more critical. Under such circumstances, adequately capturing uncertainties and understanding the potential range of inundation scenarios is crucial for life-saving decision-making, such as in evacuation and rescue operations.

In contrast, tsunami warning systems currently in operation worldwide rely primarily on a deterministic approach, outputting a single predicted value without accounting for uncertainty \cite{Ozaki_2012jdr, bernard2015evolution}.
Another significant issue with current tsunami warning systems is that, with some exceptions such as certain far-field tsunami forecasts, they provide incomplete disaster prevention information that excludes inundation---which directly causes damage---from the scope of prediction \cite{Ozaki_2012jdr, bernard2015evolution}.
While quantifying prediction uncertainty and predicting inundation are independent technical challenges, they are also intrinsically linked.
Specifically, presenting a deterministic inundation boundary line onshore under highly uncertain conditions carries the fatal risk of unintentionally conveying a misleading message that areas outside the boundary are safe. Consequently, this misrepresentation may hinder appropriate evacuation decisions by residents.

To realize real-time inundation forecasting, recent machine learning (ML)-based methods have shown great potential. High-resolution and wide-area inundation forecasting in real time has traditionally demanded massive amounts of computational resources \cite{Oishi}. However, it has been demonstrated that by training ML models to learn the nonlinear relationship between observations and inundation and employing them as surrogate models to replace computationally expensive numerical simulations, inundation predictions can be obtained instantaneously with minimal computational costs \cite{mulia2022machine, makinoshima2021deep,fauzi2020rapid, sriyanto2025rapid, MULIA2026105119}.

Despite these significant advancements in computational efficiency, the quantification of prediction uncertainty remains insufficiently studied. Although attempts have been made at ensemble-type tsunami inundation forecasting that exploit the rapid inference of ML models, the primary objectives have been the sensitivity analysis of observation errors or evaluation of wave source uncertainty, without delving into the validation of the statistical reliability of the predicted uncertainty \cite{mulia2022machine, MULIA2026105119}. Furthermore, other existing probabilistic tsunami forecasting studies limit their prediction targets to warning levels or coastal wave heights \cite{selva2021probabilistic, giles2021faster, cordrie2025dynamic, zhao2025ensemble}.

Therefore, in this study, we developed a prediction model named ``\textsf{GenTEW} (Generative Tsunami Early Warning)'' using diffusion models, a type of generative AI, providing a probabilistic real-time tsunami inundation forecast capable of quantitatively evaluating uncertainty. 
While diffusion models are a driving force behind the recent breakthrough advances in generative AI \cite{ho2020denoising, rombach2022high}, 
their probabilistic generative capabilities have also been reported to enable high-quality ensemble forecasting in the meteorological field \cite{price2024gencast}. 
We demonstrate that \textsf{GenTEW} simultaneously achieves high prediction accuracy and appropriate uncertainty quantification, not only under the current high-density observation networks in waters near Japan \cite{Mochizuki2018snet} but also under the sparse observation conditions that existed during the 2011 Tohoku-oki earthquake \cite{fujii2011tohoku}. 
This model successfully captures the large uncertainty in the early stages when observation data are lacking and dynamically refines the prediction reliability as data accumulate. 

\begin{figure*}
\centering 
\includegraphics[page=1, width=\textwidth]{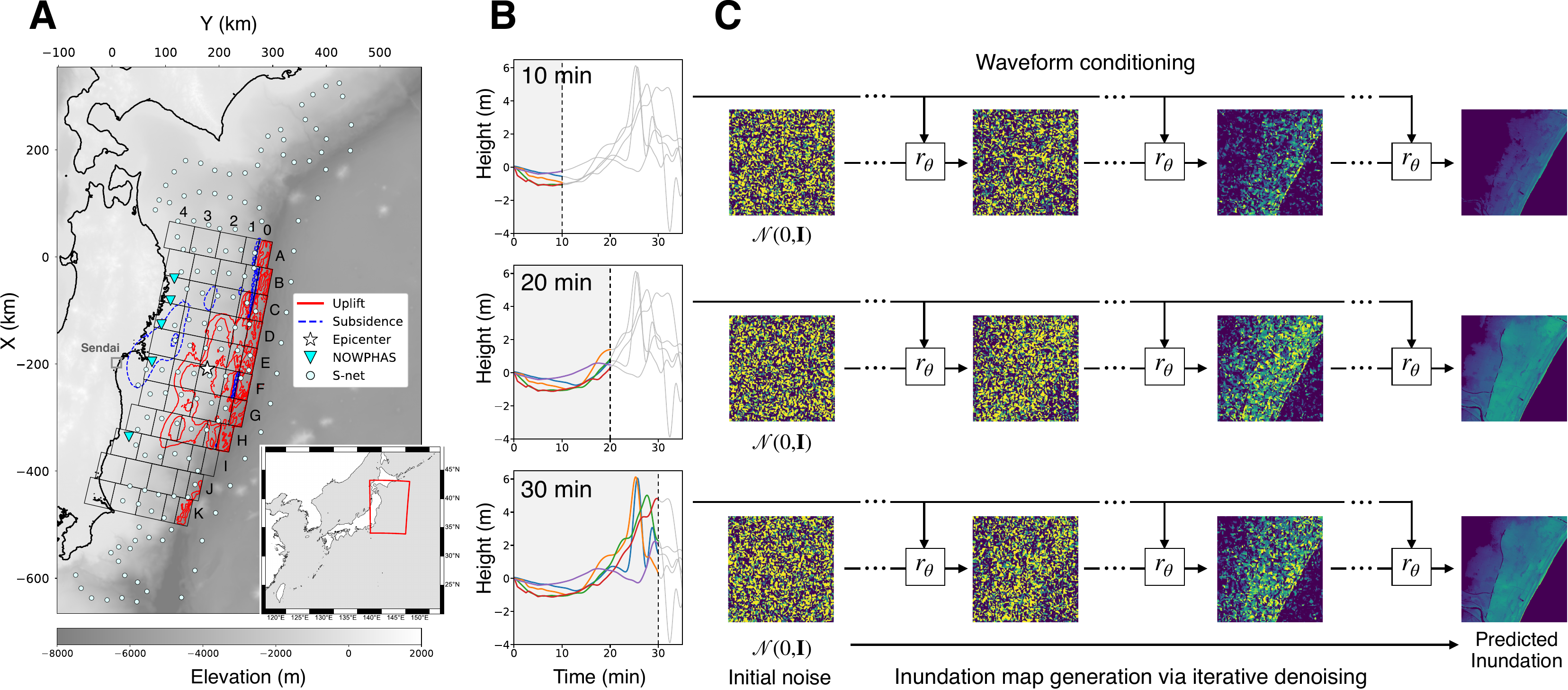} 
\caption{\textbf{Overview of real-time tsunami inundation forecasting using a conditional latent diffusion model.} \textbf{A}, Geographical setting of the target area (Pacific coast of the Tohoku region). Sea surface fluctuations of the 2011 Tohoku-oki tsunami source model estimated by Satake et al. (2013) \cite{satake2013time} are shown with contour lines at 1~m intervals (red and blue lines indicate water level rise due to seafloor uplift and decrease due to subsidence, respectively), along with the assumed subfault setting. The map also illustrates the spatial arrangement of the offshore tsunami observation networks used in this study (NOWPHAS and S-net), alongside the coastal area of Sendai city, which is the target region for inundation forecasting. \textbf{B}, Actual observed waveforms from five NOWPHAS stations recorded during the 2011 Tohoku-oki earthquake are shown \cite{kawai2012nowphas}. Water level time-series data from multiple stations, accumulated at 10, 20, and 30 minutes postearthquake, serve as conditioning guides for inferring the spatial distribution of inundation. \textbf{C}, Inundation map generation process via iterative denoising. During inference, starting from pure Gaussian noise $\mathcal{N}(0, \mathbf{I})$ in the latent space, the trained denoising model ($r_\theta$) progressively removes noise, yielding the final high-resolution inundation distribution for each elapsed time.}
\label{fig:overview} 
\end{figure*}

\section{GenTEW approach}
\textsf{GenTEW} is a ``Wave2Image''-type probabilistic tsunami prediction model that directly estimates the inland inundation distribution (maximum inundation depth or arrival time) from offshore observation waveforms, generating ensemble predictions at a 15-m resolution ($512 \times 512$ grid cells) for the target coastal area (Fig.~\ref{fig:ed_ensemble_all}A). 
This model replaces the traditional deterministic prediction approach with a latent diffusion model (LDM), a generative AI architecture that directly learns the probability distribution of the data themselves \cite{rombach2022high, ho2020denoising}. 
Figure~\ref{fig:overview} illustrates the \textsf{GenTEW} workflow; tsunami waveforms are obtained from offshore observatories (Fig.~\ref{fig:overview}A, B), and the LDM generates tsunami inundation maps through iterative denoising conditioned on the waveform features (Fig.~\ref{fig:overview}C).
To cover diverse tsunami scenarios, approximately 2,000 tsunami inundation simulation datasets generated from fault models based on probabilistic algorithms were used for training and evaluation in this paper. 

\textsf{GenTEW} models the conditional probability distribution $P(Y|W^t)$ of the 2D inundation distribution map $Y$, conditioned on the waveform time-series data $W^t$ obtained from multiple observation points up to a specific elapsed time $t$ (between 2.5 and 50 minutes in this study) after the earthquake occurrence. For $W^t$, water level waveforms from a maximum of 150 points---comprising 145 points from the Seafloor Observation Network for Earthquakes and Tsunamis along the Japan Trench (S-net) and 5 points from the Nationwide Ocean Wave Information Network for Ports and Harbours (NOWPHAS) \cite{Mochizuki2018snet, kawai2012nowphas}---are used and fed into a uniquely designed waveform encoder based on the transformer architecture \cite{vaswani2017attention}.

In the inference process, starting from pure Gaussian noise in the latent space, the inundation distribution is generated through a reverse diffusion process guided by the features extracted by the waveform encoder. By varying this initial noise, ensemble forecasting that represents the uncertainty inherent in the model becomes possible \cite{price2024gencast}.

Furthermore, in this method, we introduce perturbations to the input waveforms, inspired by the input perturbations used in meteorological ensemble forecasting \cite{Leutbecher2008}. This approach enables a comprehensive evaluation of uncertainty, including aleatoric uncertainty caused by observation errors and epistemic model uncertainty (refer to Supplementary Materials for details on noise addition).
Compared with conventional high-cost tsunami simulations, \textsf{GenTEW} is extremely fast and can generate 51 inundation ensemble members in approximately 38 seconds on a single NVIDIA GH200 GPU.

\begin{figure*}
\centering 
\includegraphics[page=2, width=\textwidth]{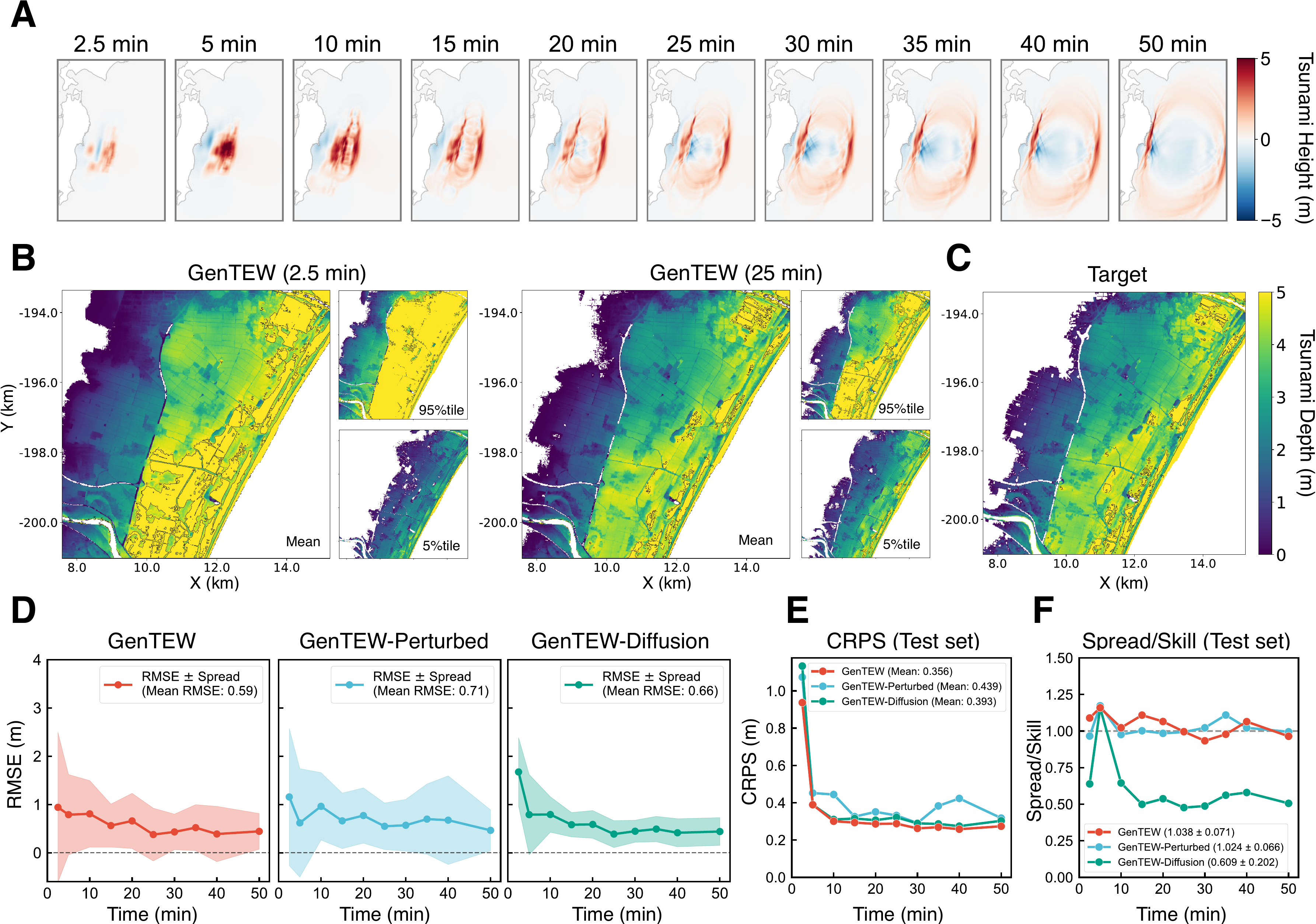} 
\caption{\textbf{Quantitative evaluation of \textsf{GenTEW} using a simulated scenario of the 2011 Tohoku-oki earthquake and test data.} Results of the ensemble inference in the 2011 scenario (synthesized S-net observation waveforms) based on the tsunami source model of Satake et al. \cite{satake2013time}, alongside a prediction accuracy evaluation using a test dataset of 200 scenarios generated via physical simulations from stochastic tsunami sources. \textbf{A}, State of tsunami propagation across the source region from 2.5 to 50 minutes post-event. \textbf{B}, Mean inundation depth map predicted by \textsf{GenTEW}, and inference results of the ensemble members corresponding to the 5th and 95th percentiles (the 3rd smallest and the 3rd largest) of total inundation volume. \textbf{C}, Ground truth inundation map generated by the target physical simulation. \textbf{D}, Time evolution of the prediction error (Root Mean Square Error, RMSE) of the ensemble mean. The shaded areas indicate the ensemble standard deviation (Spread). \textbf{E}, \textbf{F}, Average scores across the entire test dataset. \textbf{E}, CRPS \cite{price2024gencast}, indicating the overall predictive performance of the probabilistic forecast. \textbf{F}, Spread-skill ratio, indicating the reliability of uncertainty quantification. While \textsf{GenTEW-Diffusion} exhibits overconfidence, \textsf{GenTEW} hovers around the ideal value of 1.0 across all time periods, achieving appropriate calibration.}
\label{fig:ensemble_tradeoff}
\end{figure*}

\section{Baselines and evaluation protocol}
To independently evaluate the contributions of the two uncertainty generation mechanisms in \textsf{GenTEW}, we constructed two baselines as ablation models: ``\textsf{GenTEW-Diffusion},'' which relies on the inherent probabilistic nature of LDMs, and ``\textsf{GenTEW-Perturbed},'' which relies on perturbations to the input waveforms while keeping the inference process deterministic (refer to Supplementary Materials for details).

To compare these three methods, we established a two-stage evaluation protocol. First, we evaluated the models using an independent test dataset (200 scenarios) not used for training, with the inundation results from nonlinear physical simulations \cite{baba2015parallel} serving as the ground truth. Second, we performed an empirical case study using the 2011 Tohoku-oki earthquake scenario. The latter includes not only an evaluation with waveforms synthesized through simulations (Fig.~\ref{fig:ensemble_tradeoff}A) assuming the current high-density observation networks (S-net and NOWPHAS) but also a verification of robustness using historical real-world data. This verification employs the sparse actual observation waveforms recorded solely by NOWPHAS in 2011 \cite{kawai2012nowphas} (Fig.~\ref{fig:overview}B) as input and uses the actual inundation depth survey results \cite{mori2011survey} and inundation extent \cite{gsi2011tsunami} as the ground truth.

\begin{figure*}
\centering 
\includegraphics[page=3, width=\textwidth]{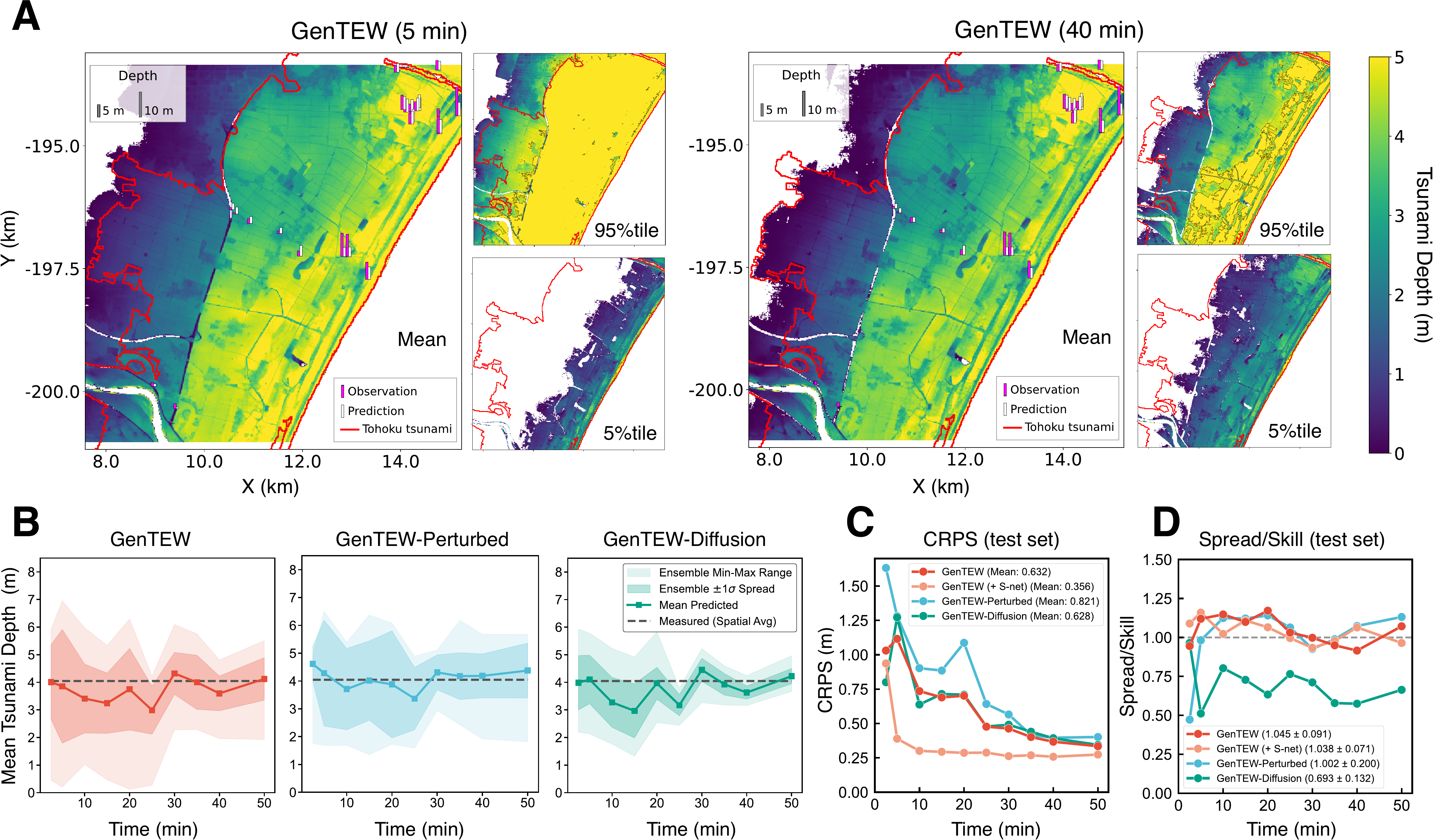} 
\caption{\textbf{Empirical evaluation of \textsf{GenTEW} using actual observational data from the 2011 Tohoku-oki earthquake.} \textbf{A}, Comparison of the predicted inundation depth distributions with the postdisaster survey results. The predicted mean maps and the 5th and 95th percentile inference results at 5 and 40 minutes post-event are shown. Actual inundation trace surveys (depths indicated by bar graphs \cite{mori2011survey} and extents, by red polygons \cite{gsi2011tsunami}) are overlaid. Predicted depths are shown for specific locations where explicit inundation depth data are provided in the survey results \cite{mori2011survey}. \textbf{B}, Comparison of the spatial means of the measured and predicted inundation depths at all survey locations shown in \textbf{A}. The shaded bands indicate the ensemble standard deviation ($\pm 1\sigma$ spread, inner band) and the range bounded by the predicted mean depths from the specific individual members that yielded the maximum and minimum values (min-max range, outer band). For instance, 5 minutes post-event, the predicted mean inundation depth of \textsf{GenTEW} across the area exhibits a wide spread ranging from approximately 0~m to 7~m, demonstrating high prediction uncertainty. \textbf{C}, \textbf{D}, Quantitative evaluation across the entire test dataset under sparse observation conditions. For reference, the results using the S-net (from Fig.~\ref{fig:ensemble_tradeoff}) are shown as faint red lines.}
\label{fig:gentew_performance_2011} 
\end{figure*}

\section{Skillful and calibrated probabilistic forecasts} 

Figure~\ref{fig:ensemble_tradeoff} shows the tsunami inundation prediction results for the coastal area of Sendai city (topography shown in Fig.~\ref{fig:ed_ensemble_all}A) under the 2011 Tohoku-oki earthquake scenario. 
These predictions are based on synthetic waveforms generated from the estimated source model (Fig.~\ref{fig:overview}A) by Satake et al. (2013) \cite{satake2013time}, assuming the currently operational high-density observation networks (S-net and NOWPHAS).
The mean map of the ensemble predictions generated by \textsf{GenTEW} (Fig.~\ref{fig:ensemble_tradeoff}B; Fig.~\ref{fig:ed_time_evolution_snet}) shows excellent spatial agreement with the target (Fig.~\ref{fig:ensemble_tradeoff}C), capturing even fine-scale topographical effects. These results confirm that the proposed generative-AI-based method achieves high fundamental performance for detailed tsunami inundation forecasting.

Furthermore, immediately after the earthquake (2.5 minutes post-event), when the fault rupture is incomplete and observation information is critically lacking, \textsf{GenTEW} outputs ensemble members ranging from scenarios where inundation remains confined to a narrow area to those causing severe inundation of more than 5~m across a wide area (Fig.~\ref{fig:ensemble_tradeoff}B; Fig.~\ref{fig:ed_ensemble_all}B). This wide variance demonstrates that \textsf{GenTEW} accurately captures the high prediction uncertainty in the initial stage. As the observed waveform data subsequently accumulate over time, the variance among the ensemble members decreases, and all the members converge with high accuracy toward the actual inundation distribution (Fig.~\ref{fig:ensemble_tradeoff}B and Fig.~\ref{fig:ed_ensemble_all}C, both at 25 minutes post-event).

In \textsf{GenTEW}, the prediction spread (standard deviation) steadily narrows as observation data accumulate (Fig.~\ref{fig:ensemble_tradeoff}D). 
Although the baseline models exhibit a similar trend, \textsf{GenTEW} demonstrates superior statistical properties with higher accuracy and a well-calibrated balance between the RMSE and spread.
The strong ability of this method to quantify uncertainty becomes even clearer when focusing on its statistical performance against simulation-generated test data. The proposed method achieves high calibration performance (spread-skill ratio of $\approx 1.0$), accurately capturing its own prediction errors (Fig.~\ref{fig:ensemble_tradeoff}F), which demonstrates the ability of \textsf{GenTEW} to properly capture dynamic changes in uncertainty and provide a highly accurate ensemble mean. Moreover, as the results of the ablation study indicate, the method relying solely on diffusion noise (\textsf{GenTEW-Diffusion}) exhibited overconfidence, and the method relying solely on input perturbations (\textsf{GenTEW-Perturbed}) decreased fundamental accuracy. In contrast, \textsf{GenTEW} overcame the trade-off between the two, achieving both the highest accuracy and calibration (Fig.~\ref{fig:ensemble_tradeoff}F), resulting in the best comprehensive performance measured by the continuous ranked probability score (CRPS) (Fig.~\ref{fig:ensemble_tradeoff}E). 

The results of this comparative experiment reveal the importance of an appropriate partitioning of roles for uncertainty in ensemble generation. \textsf{GenTEW-Perturbed} applied excessive perturbations in an attempt to represent not only observation errors but also the model's inherent uncertainty exclusively through input disturbances; as a result, the physical consistency of the input waveforms was compromised, sacrificing prediction accuracy. In contrast, in \textsf{GenTEW}, aleatoric uncertainty caused by observation noise was represented as additive noise to the input waveforms, similar to \textsf{GenTEW-Perturbed}. For epistemic uncertainty, while relying primarily on the latent noise of the diffusion model, minimal multiplicative noise to the input waveforms was introduced as an auxiliary control variable to optimize its calibration. Consequently, optimal calibration was achieved with a smaller input noise level compared to \textsf{GenTEW-Perturbed} (Table~\ref{tab:optimization_results}). Through this appropriate partitioning of roles for uncertainty, \textsf{GenTEW} achieved ideal reliability without compromising prediction accuracy (refer to Supplementary Materials for details on noise level optimization).

Notably, under high-density observation networks including S-net, the predictive performance (CRPS) sharply improves 10 minutes post-event (Fig.~\ref{fig:ensemble_tradeoff}E). In our fault models, the maximum duration of crustal deformation is set to 5 minutes; thus, at the 10-minute point, the tsunami has had approximately 5 minutes to propagate since its full generation. Moreover, in deep ocean areas with water depths greater than 1,000~m, where approximately 80\% of S-net observation points are deployed \cite{bosai_snet_info}, a tsunami propagates more than 30~km in 5 minutes. This distance matches the average installation interval of S-net \cite{tanioka2020improvement}. Given the principles of ocean bottom pressure gauges (OBPGs), it is difficult to immediately observe the initial sea surface fluctuations accompanying the progression of fault rupture \cite{tsushima2012tsunami}. 
Therefore, the characteristic 10-minute threshold---at which the inference uncertainty of \textsf{GenTEW} rapidly declines---is consistent with the physical time required for the tsunami to reach and be captured by adjacent observation points. 
This finding strongly supports the high observation potential of the S-net network and demonstrates the effectiveness of the proposed method in enhancing its utility. 

Furthermore, we confirmed that \textsf{GenTEW} can predict the tsunami arrival time with high accuracy, similar to its predictions for inundation depth (Figs.~\ref{fig:ed_arrival_time_stats}A--E, \ref{fig:ed_time_evolution_arrival}).

\begin{figure*}
\centering 
\includegraphics[page=4, width=0.8\textwidth]{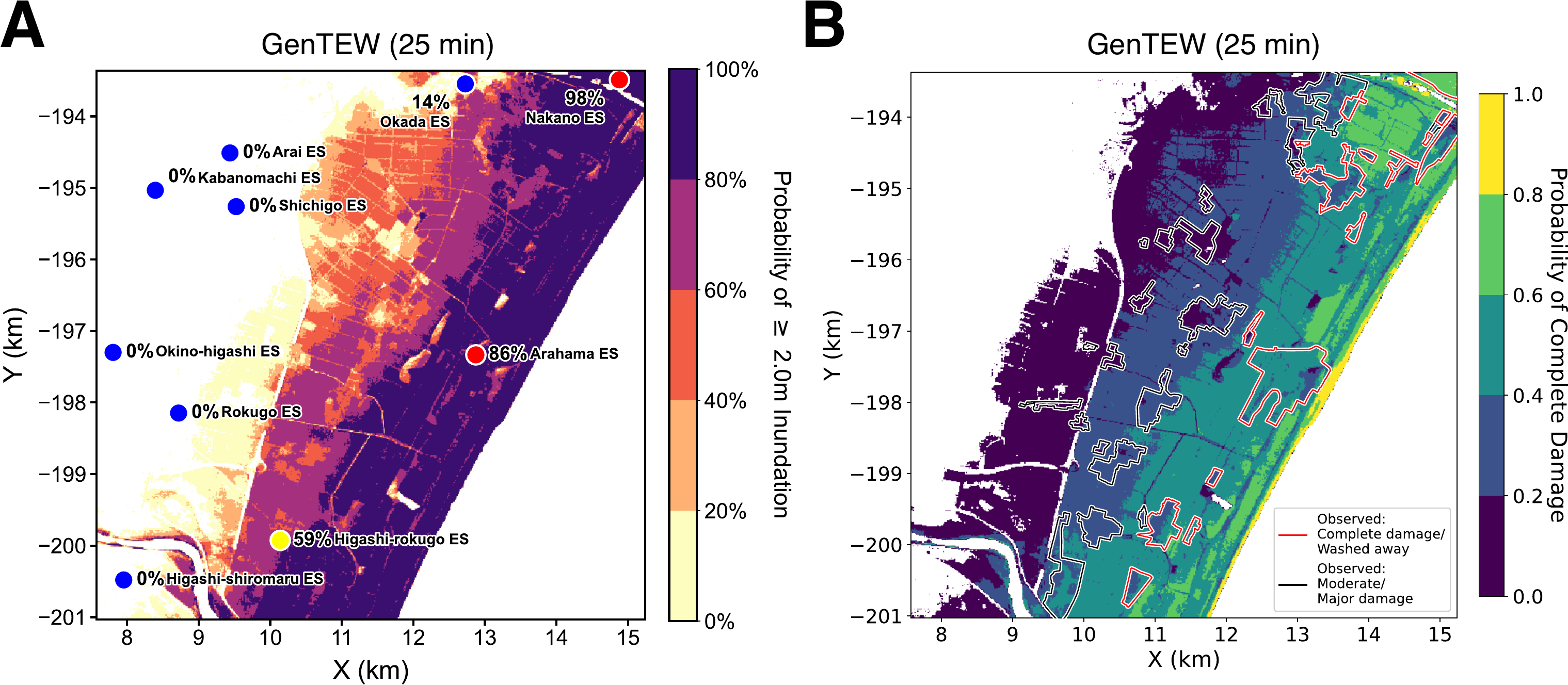} 
\caption{\textbf{Examples of probabilistic tsunami inundation forecasting products.} \textbf{A} and \textbf{B} represent an empirical use case at 25 minutes post-event using NOWPHAS observation data from 2011. \textbf{A}, Probabilistic inundation forecasting. The inundation risk ($\ge 2~\text{m}$) for each elementary school (ES) is presented as a probability. High probabilities are accurately calculated for three schools that actually suffered severe damage (Nakano ES: 98\%, Arahama ES: 86\%, Higashi-rokugo ES: 59\%). Okada ES, which did not experience severe inundation, also exhibits a nonzero probability (14\%), demonstrating that the model provides continuous probabilistic forecasts rather than binary predictions of occurrence. \textbf{B}, Real-time building damage prediction. The spatial distribution of the probability of building collapse, which is calculated by coupling the inundation probabilities from \textbf{A} with fragility curves \cite{suppasri2013building}, is shown. Compared with actual survey data of damaged building areas \cite{Sekimoto2013}, the damage probability is high (average 44.6\%) in areas where buildings were ``completely destroyed or washed away'' (red lines), and lower (average 26.0\%) in areas with ``moderate to large-scale damage'' (black lines), showing good agreement with the spatial distribution of actual damage scales. (Although high damage probabilities are also predicted outside these survey polygons, limited actual damage is reported primarily due to the absence of buildings.)}
\label{fig:rev_conf_matrix} 
\end{figure*}

\section{Robustness under sparse historical observations}
To evaluate the real-world applicability of this system, we conducted an empirical validation using historical observations. Specifically, we utilized the actual tsunami observation waveforms recorded solely by NOWPHAS \cite{kawai2012nowphas} during the 2011 Tohoku-oki tsunami (Fig.~\ref{fig:overview}B)---when S-net had not yet been deployed---as input. This validation also serves as a crucial reference for the global deployment of this method to regions where high-density observation networks are not established.

As Figures~\ref{fig:gentew_performance_2011}A and B demonstrate, the ensemble outputs steadily converge toward predictions that are consistent with the inundation depth survey results \cite{mori2011survey} and actual inundation extents \cite{gsi2011tsunami}. These results suggest that \textsf{GenTEW} possesses inherent robustness even in sparse observation environments.
However, when relying exclusively on NOWPHAS stations, which are deployed approximately 10--20~km from the coast, it took approximately 30--40 minutes for the data to accumulate and the ensemble prediction diversity to converge (Fig.~\ref{fig:gentew_performance_2011}B, C and Fig.~\ref{fig:ed_time_evolution_nowphas}). Nevertheless, a vital lead time is still obtained before the tsunami reaches the coastal area of Sendai city approximately 60 minutes after the earthquake. 

Most notably, even under such limited observation conditions, \textsf{GenTEW} appropriately quantified its own prediction uncertainty and consistently provided highly reliable probabilistic forecasts. Although its predictive performance (CRPS) is inherently lower than that of the high-density observation case with S-net, the spread-skill ratio generally remained near 1.0, maintaining the ideal calibration observed in the dense observation case (Fig.~\ref{fig:gentew_performance_2011}C, D).

\begin{figure*}
\centering 
\includegraphics[page=5, width=0.8\textwidth]{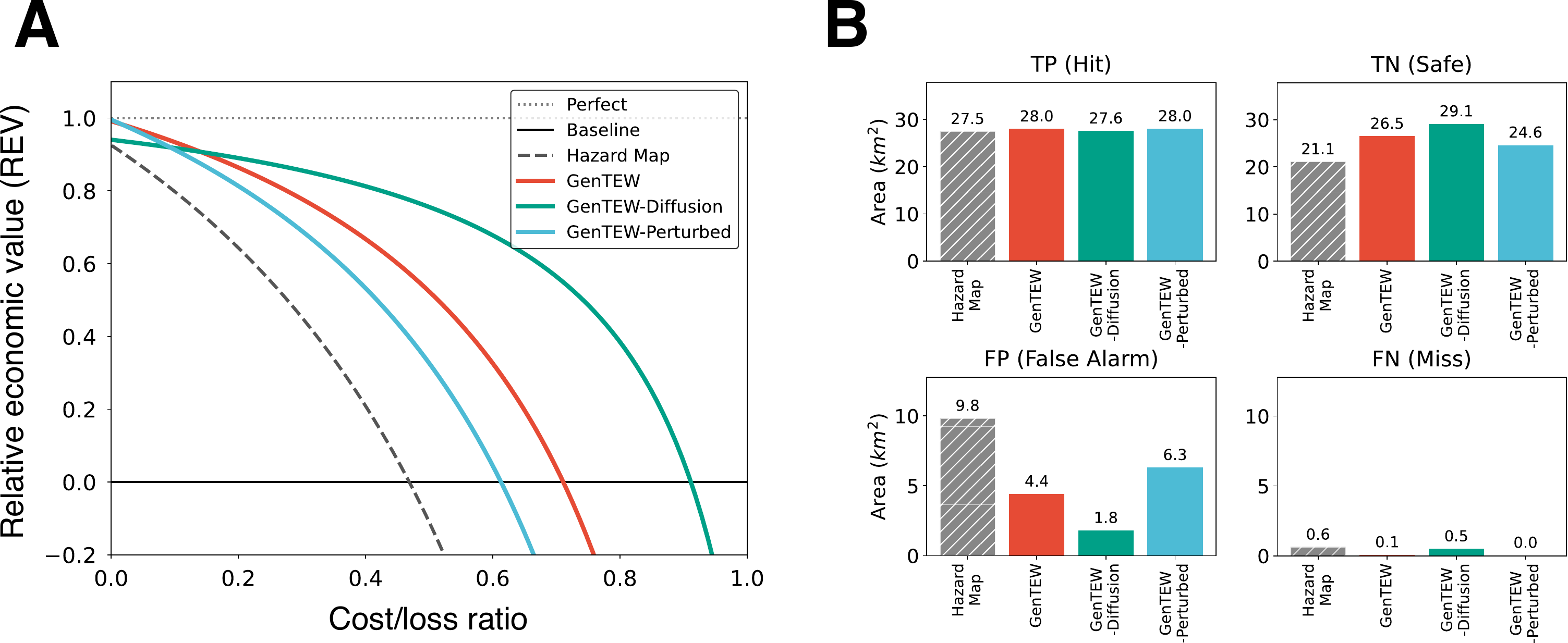} 
\caption{\textbf{Relative economic value and confusion matrices of forecasts.} \textbf{A} and \textbf{B} represent quantitative evaluations at 25 minutes post-event over the entire test dataset with NOWPHAS and S-net observations. \textbf{A}, Relative economic value (REV) curves. Decision-making based on the mean of the training data is used as the baseline. ``Perfect'' indicates an ideal prediction without any false positives (FP) or false negatives (FN). The actual inundation area of the 2011 Tohoku-oki tsunami is assumed as the hazard map. \textbf{B}, Mean area comparison based on confusion matrices, consisting of true positive (TP), true negative (TN), FP, and FN forecasts.}
\label{fig:rev} 
\end{figure*}

\section{Actionable risk assessment and economic value}
The core significance of probabilistic tsunami inundation forecasting is demonstrated by its capacity to improve the quality of evacuation decision-making. In contrast to the traditional deterministic boundary between inundated and noninundated areas, \textsf{GenTEW} mitigates binary cognitive biases by presenting the inundation probability at specific inland locations (such as schools) as a continuous gradient, much like the precipitation probability (Fig.~\ref{fig:rev_conf_matrix}A). Furthermore, by integrating the inferred probability distribution of the inundation depth with fragility curves \cite{suppasri2013building, koshimura2009tsunami}, the statistical risks to buildings or human safety can be directly evaluated even in real time (refer to Supplementary Materials for details). When it is applied to the 2011 Tohoku-oki tsunami, these evaluation results closely match the spatial distribution of actual building damage survey polygons \cite{Sekimoto2013} (Fig.~\ref{fig:rev_conf_matrix}B).

The effectiveness of \textsf{GenTEW} in supporting evacuation decisions is also corroborated by the quantitative evaluation of relative economic value (REV) \cite{richardson2000skill, price2024gencast} (refer to Supplementary Materials for details). 
A hazard map based on maximum-class assumptions derived from a static single scenario is prone to generating excessive warnings (false positives) and cannot completely avoid unexpected misses (false negatives), yielding limited economic value (Fig.~\ref{fig:rev}A, B). 
In addition, the baseline method with inferior calibration (\textsf{GenTEW-Diffusion}) suffers from a critical limitation: its value slightly diminishes in the low cost/loss ratio regime, which is directly linked to the loss of life in the tsunami damage mitigation case. 
In contrast, \textsf{GenTEW} substantially reduces missed warnings while also lowering the socioeconomic costs of excessive warnings and maintaining consistently high REV across all cost/loss regimes, showing its capability to achieve both high reliability to protect human lives and socioeconomic rationality.

\section{Conclusion}
This study introduced ``\textsf{GenTEW},'' a generative AI tsunami forecasting system that probabilistically infers inland inundation distributions from offshore observation waveforms using a conditional LDM. Through validation, \textsf{GenTEW} fulfilled three critical requirements for probabilistic tsunami forecasting. First, it achieved highly calibrated predictions (spread-skill ratio $\approx 1.0$) without compromising the baseline accuracy. Second, it demonstrated inherent robustness even under limited observation conditions, including short observation times and sparse sensor networks. Third, it significantly overcame the computational bottleneck of conventional nonlinear tsunami simulations, realizing an operationally viable inference speed (approximately 38 seconds).

Since the inference performance of this method inherently depends on the distribution of the training data, expanding the infrastructure for generating training data remains a future challenge for full-scale societal deployment. In particular, to ensure the system's robustness against potential ``unexpected'' megathrust earthquakes in the future, it is essential to prepare a diverse and comprehensive set of fault source scenarios. Furthermore, to support decision-making at a more local scale, such as at the level of individual coastal structures, further advancement and higher resolution of the nonlinear physical simulations underlying data generation are required. 

The shift from deterministic to probabilistic forecasting presented in this study enhances the practical feasibility of inundation forecasting by avoiding the risks associated with definitive deterministic predictions. 
Additionally, we demonstrated that by integrating the probabilistic hazard predictions from \textsf{GenTEW} with fragility curves, the tsunami early warning can be transformed from macrolevel information, such as ``wave height at the coastline,'' into intuitive and realistic socioeconomic risks, such as ``the probability of specific onshore facilities being damaged.''

\section*{Data availability}
Tsunami observation records were provided by NOWPHAS (MLIT and PARI) \cite{kawai2012nowphas}. Inundation heights were obtained from the 2011 Tohoku Earthquake Tsunami Joint Survey (TTJS) Group (\url{https://www.coastal.jp/ttjt/}) \cite{mori2011survey}. Building damage data and inundation area maps were obtained from the Great East Japan Earthquake Recovery Support Survey Archive (\url{http://fukkou.csis.u-tokyo.ac.jp/}) \cite{Sekimoto2013} and the Geospatial Information Authority of Japan (GSI) \cite{gsi2011tsunami}, respectively. The terrain model integrates the Global tsunami Terrain Model (GtTM) \cite{chikasada2020gttm} and coastal topography data from the Tohoku Regional Development Bureau, MLIT.

\section*{Acknowledgments}
The computational results were obtained using the Wisteria/BDEC-01 and Miyabi supercomputer systems at the Information Technology Center, University of Tokyo. 
This work was supported by JSPS KAKENHI (Grant Numbers JP24K01128 and JP23K23018).

\bibliography{Oishi_et_al}

\begin{thebibliography}{51}%
\makeatletter
\providecommand \@ifxundefined [1]{%
 \@ifx{#1\undefined}
}%
\providecommand \@ifnum [1]{%
 \ifnum #1\expandafter \@firstoftwo
 \else \expandafter \@secondoftwo
 \fi
}%
\providecommand \@ifx [1]{%
 \ifx #1\expandafter \@firstoftwo
 \else \expandafter \@secondoftwo
 \fi
}%
\providecommand \natexlab [1]{#1}%
\providecommand \enquote  [1]{``#1''}%
\providecommand \bibnamefont  [1]{#1}%
\providecommand \bibfnamefont [1]{#1}%
\providecommand \citenamefont [1]{#1}%
\providecommand \href@noop [0]{\@secondoftwo}%
\providecommand \href [0]{\begingroup \@sanitize@url \@href}%
\providecommand \@href[1]{\@@startlink{#1}\@@href}%
\providecommand \@@href[1]{\endgroup#1\@@endlink}%
\providecommand \@sanitize@url [0]{\catcode `\\12\catcode `\$12\catcode `\&12\catcode `\#12\catcode `\^12\catcode `\_12\catcode `\%12\relax}%
\providecommand \@@startlink[1]{}%
\providecommand \@@endlink[0]{}%
\providecommand \url  [0]{\begingroup\@sanitize@url \@url }%
\providecommand \@url [1]{\endgroup\@href {#1}{\urlprefix }}%
\providecommand \urlprefix  [0]{URL }%
\providecommand \Eprint [0]{\href }%
\providecommand \doibase [0]{https://doi.org/}%
\providecommand \selectlanguage [0]{\@gobble}%
\providecommand \bibinfo  [0]{\@secondoftwo}%
\providecommand \bibfield  [0]{\@secondoftwo}%
\providecommand \translation [1]{[#1]}%
\providecommand \BibitemOpen [0]{}%
\providecommand \bibitemStop [0]{}%
\providecommand \bibitemNoStop [0]{.\EOS\space}%
\providecommand \EOS [0]{\spacefactor3000\relax}%
\providecommand \BibitemShut  [1]{\csname bibitem#1\endcsname}%
\let\auto@bib@innerbib\@empty
\bibitem [{\citenamefont {Neumann}\ \emph {et~al.}(2015)\citenamefont {Neumann}, \citenamefont {Vafeidis}, \citenamefont {Zimmermann},\ and\ \citenamefont {Nicholls}}]{neumann2015future}%
  \BibitemOpen
  \bibfield  {author} {\bibinfo {author} {\bibfnamefont {B.}~\bibnamefont {Neumann}}, \bibinfo {author} {\bibfnamefont {A.~T.}\ \bibnamefont {Vafeidis}}, \bibinfo {author} {\bibfnamefont {J.}~\bibnamefont {Zimmermann}},\ and\ \bibinfo {author} {\bibfnamefont {R.~J.}\ \bibnamefont {Nicholls}},\ }\bibfield  {title} {\bibinfo {title} {Future coastal population growth and exposure to sea-level rise and coastal flooding-a global assessment},\ }\href@noop {} {\bibfield  {journal} {\bibinfo  {journal} {PLoS ONE}\ }\textbf {\bibinfo {volume} {10}},\ \bibinfo {pages} {e0118571} (\bibinfo {year} {2015})}\BibitemShut {NoStop}%
\bibitem [{\citenamefont {Synolakis}\ and\ \citenamefont {Bernard}(2006)}]{synolakis2006tsunami}%
  \BibitemOpen
  \bibfield  {author} {\bibinfo {author} {\bibfnamefont {C.~E.}\ \bibnamefont {Synolakis}}\ and\ \bibinfo {author} {\bibfnamefont {E.~N.}\ \bibnamefont {Bernard}},\ }\bibfield  {title} {\bibinfo {title} {Tsunami science before and beyond {Boxing Day} 2004},\ }\href {https://doi.org/10.1098/rsta.2006.1824} {\bibfield  {journal} {\bibinfo  {journal} {Philos. Trans. R. Soc. A}\ }\textbf {\bibinfo {volume} {364}},\ \bibinfo {pages} {2231} (\bibinfo {year} {2006})}\BibitemShut {NoStop}%
\bibitem [{\citenamefont {Angove}\ \emph {et~al.}(2019)\citenamefont {Angove} \emph {et~al.}}]{angove2019ocean}%
  \BibitemOpen
  \bibfield  {author} {\bibinfo {author} {\bibfnamefont {M.}~\bibnamefont {Angove}} \emph {et~al.},\ }\bibfield  {title} {\bibinfo {title} {Ocean observations required to minimize uncertainty in global tsunami forecasts, warnings, and emergency response},\ }\href {https://doi.org/10.3389/fmars.2019.00350} {\bibfield  {journal} {\bibinfo  {journal} {Front. Mar. Sci.}\ }\textbf {\bibinfo {volume} {6}},\ \bibinfo {pages} {350} (\bibinfo {year} {2019})}\BibitemShut {NoStop}%
\bibitem [{\citenamefont {Goda}\ and\ \citenamefont {Song}(2016)}]{goda2016uncertainty}%
  \BibitemOpen
  \bibfield  {author} {\bibinfo {author} {\bibfnamefont {K.}~\bibnamefont {Goda}}\ and\ \bibinfo {author} {\bibfnamefont {J.}~\bibnamefont {Song}},\ }\bibfield  {title} {\bibinfo {title} {Uncertainty modeling and visualization for tsunami hazard and risk mapping: a case study for the 2011 tohoku earthquake},\ }\href {https://doi.org/10.1007/s00477-015-1146-x} {\bibfield  {journal} {\bibinfo  {journal} {Stoch. Environ. Res. Risk Assess.}\ }\textbf {\bibinfo {volume} {30}},\ \bibinfo {pages} {2271} (\bibinfo {year} {2016})}\BibitemShut {NoStop}%
\bibitem [{\citenamefont {Selva}\ \emph {et~al.}(2021)\citenamefont {Selva} \emph {et~al.}}]{selva2021probabilistic}%
  \BibitemOpen
  \bibfield  {author} {\bibinfo {author} {\bibfnamefont {J.}~\bibnamefont {Selva}} \emph {et~al.},\ }\bibfield  {title} {\bibinfo {title} {Probabilistic tsunami forecasting for early warning},\ }\href {https://doi.org/10.1038/s41467-021-25815-w} {\bibfield  {journal} {\bibinfo  {journal} {Nat. Commun.}\ }\textbf {\bibinfo {volume} {12}},\ \bibinfo {pages} {5677} (\bibinfo {year} {2021})}\BibitemShut {NoStop}%
\bibitem [{\citenamefont {Grezio}\ \emph {et~al.}(2017)\citenamefont {Grezio} \emph {et~al.}}]{grezio2017probabilistic}%
  \BibitemOpen
  \bibfield  {author} {\bibinfo {author} {\bibfnamefont {A.}~\bibnamefont {Grezio}} \emph {et~al.},\ }\bibfield  {title} {\bibinfo {title} {Probabilistic tsunami hazard analysis: multiple sources and global applications},\ }\href {https://doi.org/10.1002/2017RG000579} {\bibfield  {journal} {\bibinfo  {journal} {Rev. Geophys.}\ }\textbf {\bibinfo {volume} {55}},\ \bibinfo {pages} {1158} (\bibinfo {year} {2017})}\BibitemShut {NoStop}%
\bibitem [{\citenamefont {Titov}\ and\ \citenamefont {Gonzalez}(1997)}]{titov1997implementation}%
  \BibitemOpen
  \bibfield  {author} {\bibinfo {author} {\bibfnamefont {V.~V.}\ \bibnamefont {Titov}}\ and\ \bibinfo {author} {\bibfnamefont {F.~I.}\ \bibnamefont {Gonzalez}},\ }\bibfield  {title} {\bibinfo {title} {Implementation and testing of the method of splitting tsunami ({MOST}) model},\ }\href {https://repository.library.noaa.gov/view/noaa/10979} {\  (\bibinfo {year} {1997})},\ \bibinfo {note} {{NOAA technical memorandum ERL PMEL, 112, Contribution (Pacific Marine Environmental Laboratory (U.S.)), no. 1927}}\BibitemShut {NoStop}%
\bibitem [{\citenamefont {Ozaki}(2012)}]{Ozaki_2012jdr}%
  \BibitemOpen
  \bibfield  {author} {\bibinfo {author} {\bibfnamefont {T.}~\bibnamefont {Ozaki}},\ }\bibfield  {title} {\bibinfo {title} {{JMA's} tsunami warning for the 2011 {Great Tohoku} earthquake and tsunami warning improvement plan},\ }\href {https://doi.org/10.20965/jdr.2012.p0439} {\bibfield  {journal} {\bibinfo  {journal} {J. Disaster Res.}\ }\textbf {\bibinfo {volume} {7}},\ \bibinfo {pages} {439} (\bibinfo {year} {2012})}\BibitemShut {NoStop}%
\bibitem [{\citenamefont {Bernard}\ and\ \citenamefont {Titov}(2015)}]{bernard2015evolution}%
  \BibitemOpen
  \bibfield  {author} {\bibinfo {author} {\bibfnamefont {E.}~\bibnamefont {Bernard}}\ and\ \bibinfo {author} {\bibfnamefont {V.}~\bibnamefont {Titov}},\ }\bibfield  {title} {\bibinfo {title} {Evolution of tsunami warning systems and products},\ }\href {https://doi.org/10.1098/rsta.2014.0371} {\bibfield  {journal} {\bibinfo  {journal} {Philos. Trans. R. Soc. A}\ }\textbf {\bibinfo {volume} {373}},\ \bibinfo {pages} {20140371} (\bibinfo {year} {2015})}\BibitemShut {NoStop}%
\bibitem [{\citenamefont {Oishi}\ \emph {et~al.}(2015)\citenamefont {Oishi}, \citenamefont {Imamura},\ and\ \citenamefont {Sugawara}}]{Oishi}%
  \BibitemOpen
  \bibfield  {author} {\bibinfo {author} {\bibfnamefont {Y.}~\bibnamefont {Oishi}}, \bibinfo {author} {\bibfnamefont {F.}~\bibnamefont {Imamura}},\ and\ \bibinfo {author} {\bibfnamefont {D.}~\bibnamefont {Sugawara}},\ }\bibfield  {title} {\bibinfo {title} {Near-field tsunami inundation forecast using the parallel {TUNAMI-N2} model: application to the 2011 {Tohoku-Oki} earthquake combined with source inversions},\ }\href {https://doi.org/10.1002/2014GL062577} {\bibfield  {journal} {\bibinfo  {journal} {Geophys. Res. Lett.}\ }\textbf {\bibinfo {volume} {42}},\ \bibinfo {pages} {1083} (\bibinfo {year} {2015})}\BibitemShut {NoStop}%
\bibitem [{\citenamefont {Mulia}\ \emph {et~al.}(2022)\citenamefont {Mulia}, \citenamefont {Ueda}, \citenamefont {Miyoshi}, \citenamefont {Gusman},\ and\ \citenamefont {Satake}}]{mulia2022machine}%
  \BibitemOpen
  \bibfield  {author} {\bibinfo {author} {\bibfnamefont {I.~E.}\ \bibnamefont {Mulia}}, \bibinfo {author} {\bibfnamefont {N.}~\bibnamefont {Ueda}}, \bibinfo {author} {\bibfnamefont {T.}~\bibnamefont {Miyoshi}}, \bibinfo {author} {\bibfnamefont {A.~R.}\ \bibnamefont {Gusman}},\ and\ \bibinfo {author} {\bibfnamefont {K.}~\bibnamefont {Satake}},\ }\bibfield  {title} {\bibinfo {title} {Machine learning-based tsunami inundation prediction derived from offshore observations},\ }\href {https://doi.org/10.1038/s41467-022-33253-5} {\bibfield  {journal} {\bibinfo  {journal} {Nat. Commun.}\ }\textbf {\bibinfo {volume} {13}},\ \bibinfo {pages} {5489} (\bibinfo {year} {2022})}\BibitemShut {NoStop}%
\bibitem [{\citenamefont {Makinoshima}\ \emph {et~al.}(2021)\citenamefont {Makinoshima}, \citenamefont {Oishi}, \citenamefont {Yamazaki}, \citenamefont {Furumura},\ and\ \citenamefont {Imamura}}]{makinoshima2021deep}%
  \BibitemOpen
  \bibfield  {author} {\bibinfo {author} {\bibfnamefont {F.}~\bibnamefont {Makinoshima}}, \bibinfo {author} {\bibfnamefont {Y.}~\bibnamefont {Oishi}}, \bibinfo {author} {\bibfnamefont {T.}~\bibnamefont {Yamazaki}}, \bibinfo {author} {\bibfnamefont {T.}~\bibnamefont {Furumura}},\ and\ \bibinfo {author} {\bibfnamefont {F.}~\bibnamefont {Imamura}},\ }\bibfield  {title} {\bibinfo {title} {Early forecasting of tsunami inundation from tsunami and geodetic observation data with convolutional neural networks},\ }\href {https://doi.org/10.1038/s41467-021-22348-0} {\bibfield  {journal} {\bibinfo  {journal} {Nat. Commun.}\ }\textbf {\bibinfo {volume} {12}},\ \bibinfo {pages} {2253} (\bibinfo {year} {2021})}\BibitemShut {NoStop}%
\bibitem [{\citenamefont {Fauzi}\ and\ \citenamefont {Mizutani}(2020)}]{fauzi2020rapid}%
  \BibitemOpen
  \bibfield  {author} {\bibinfo {author} {\bibfnamefont {A.}~\bibnamefont {Fauzi}}\ and\ \bibinfo {author} {\bibfnamefont {N.}~\bibnamefont {Mizutani}},\ }\bibfield  {title} {\bibinfo {title} {Machine learning algorithms for real-time tsunami inundation forecasting: a case study in {Nankai} region},\ }\href {https://doi.org/10.1007/s00024-019-02364-4} {\bibfield  {journal} {\bibinfo  {journal} {Pure Appl. Geophys.}\ }\textbf {\bibinfo {volume} {177}},\ \bibinfo {pages} {1437} (\bibinfo {year} {2020})}\BibitemShut {NoStop}%
\bibitem [{\citenamefont {Sriyanto}\ \emph {et~al.}(2025)\citenamefont {Sriyanto}, \citenamefont {Adriano}, \citenamefont {Fujii},\ and\ \citenamefont {Koshimura}}]{sriyanto2025rapid}%
  \BibitemOpen
  \bibfield  {author} {\bibinfo {author} {\bibfnamefont {S.~P.~D.}\ \bibnamefont {Sriyanto}}, \bibinfo {author} {\bibfnamefont {B.}~\bibnamefont {Adriano}}, \bibinfo {author} {\bibfnamefont {Y.}~\bibnamefont {Fujii}},\ and\ \bibinfo {author} {\bibfnamefont {S.}~\bibnamefont {Koshimura}},\ }\bibfield  {title} {\bibinfo {title} {Estimation of high-resolution tsunami inundation depth using deep learning models: case study of {Pangandaran}, {Indonesia}},\ }\href {https://doi.org/10.1016/j.oceaneng.2025.121019} {\bibfield  {journal} {\bibinfo  {journal} {Ocean Eng.}\ }\textbf {\bibinfo {volume} {330}},\ \bibinfo {pages} {121019} (\bibinfo {year} {2025})}\BibitemShut {NoStop}%
\bibitem [{\citenamefont {Mulia}\ \emph {et~al.}(2026)\citenamefont {Mulia} \emph {et~al.}}]{MULIA2026105119}%
  \BibitemOpen
  \bibfield  {author} {\bibinfo {author} {\bibfnamefont {I.~E.}\ \bibnamefont {Mulia}} \emph {et~al.},\ }\bibfield  {title} {\bibinfo {title} {Ai-based ensemble tsunami inundation forecasting},\ }\href {https://doi.org/10.1016/j.coastaleng.2026.105119} {\bibfield  {journal} {\bibinfo  {journal} {Coast. Eng.}\ }\textbf {\bibinfo {volume} {212}},\ \bibinfo {pages} {105119} (\bibinfo {year} {2026})}\BibitemShut {NoStop}%
\bibitem [{\citenamefont {Giles}\ \emph {et~al.}(2021)\citenamefont {Giles}, \citenamefont {Gopinathan}, \citenamefont {Guillas},\ and\ \citenamefont {Dias}}]{giles2021faster}%
  \BibitemOpen
  \bibfield  {author} {\bibinfo {author} {\bibfnamefont {D.}~\bibnamefont {Giles}}, \bibinfo {author} {\bibfnamefont {D.}~\bibnamefont {Gopinathan}}, \bibinfo {author} {\bibfnamefont {S.}~\bibnamefont {Guillas}},\ and\ \bibinfo {author} {\bibfnamefont {F.}~\bibnamefont {Dias}},\ }\bibfield  {title} {\bibinfo {title} {Faster than real time tsunami warning with associated hazard uncertainties},\ }\bibfield  {journal} {\bibinfo  {journal} {Front. Earth Sci.}\ }\textbf {\bibinfo {volume} {8}},\ \href {https://doi.org/10.3389/feart.2020.597865} {10.3389/feart.2020.597865} (\bibinfo {year} {2021})\BibitemShut {NoStop}%
\bibitem [{\citenamefont {Cordrie}\ \emph {et~al.}(2025)\citenamefont {Cordrie} \emph {et~al.}}]{cordrie2025dynamic}%
  \BibitemOpen
  \bibfield  {author} {\bibinfo {author} {\bibfnamefont {L.}~\bibnamefont {Cordrie}} \emph {et~al.},\ }\bibfield  {title} {\bibinfo {title} {Dynamic management of uncertainty in rapid tsunami forecasting},\ }\href {https://doi.org/10.1038/s43247-025-02586-6} {\bibfield  {journal} {\bibinfo  {journal} {Commun. Earth Environ.}\ }\textbf {\bibinfo {volume} {6}},\ \bibinfo {pages} {637} (\bibinfo {year} {2025})}\BibitemShut {NoStop}%
\bibitem [{\citenamefont {Zhao}\ and\ \citenamefont {Niu}(2025)}]{zhao2025ensemble}%
  \BibitemOpen
  \bibfield  {author} {\bibinfo {author} {\bibfnamefont {G.}~\bibnamefont {Zhao}}\ and\ \bibinfo {author} {\bibfnamefont {X.}~\bibnamefont {Niu}},\ }\bibfield  {title} {\bibinfo {title} {An ensemble forecasting method for tsunami warning},\ }\href {https://doi.org/10.1007/s11069-024-07068-0} {\bibfield  {journal} {\bibinfo  {journal} {Nat. Hazards}\ }\textbf {\bibinfo {volume} {121}},\ \bibinfo {pages} {6651} (\bibinfo {year} {2025})}\BibitemShut {NoStop}%
\bibitem [{\citenamefont {Ho}\ \emph {et~al.}(2020)\citenamefont {Ho}, \citenamefont {Jain},\ and\ \citenamefont {Abbeel}}]{ho2020denoising}%
  \BibitemOpen
  \bibfield  {author} {\bibinfo {author} {\bibfnamefont {J.}~\bibnamefont {Ho}}, \bibinfo {author} {\bibfnamefont {A.}~\bibnamefont {Jain}},\ and\ \bibinfo {author} {\bibfnamefont {P.}~\bibnamefont {Abbeel}},\ }\bibfield  {title} {\bibinfo {title} {Denoising diffusion probabilistic models},\ }in\ \href@noop {} {\emph {\bibinfo {booktitle} {Adv. Neural Inf. Process. Syst.}}},\ Vol.~\bibinfo {volume} {33}\ (\bibinfo {year} {2020})\ pp.\ \bibinfo {pages} {6840--6851}\BibitemShut {NoStop}%
\bibitem [{\citenamefont {Rombach}\ \emph {et~al.}(2022)\citenamefont {Rombach}, \citenamefont {Blattmann}, \citenamefont {Lorenz}, \citenamefont {Esser},\ and\ \citenamefont {Ommer}}]{rombach2022high}%
  \BibitemOpen
  \bibfield  {author} {\bibinfo {author} {\bibfnamefont {R.}~\bibnamefont {Rombach}}, \bibinfo {author} {\bibfnamefont {A.}~\bibnamefont {Blattmann}}, \bibinfo {author} {\bibfnamefont {D.}~\bibnamefont {Lorenz}}, \bibinfo {author} {\bibfnamefont {P.}~\bibnamefont {Esser}},\ and\ \bibinfo {author} {\bibfnamefont {B.}~\bibnamefont {Ommer}},\ }\bibfield  {title} {\bibinfo {title} {High-resolution image synthesis with latent diffusion models},\ }in\ \href@noop {} {\emph {\bibinfo {booktitle} {Proc. IEEE/CVF Conf. Comput. Vis. Pattern Recognit. (CVPR)}}}\ (\bibinfo {year} {2022})\ pp.\ \bibinfo {pages} {10684--10695}\BibitemShut {NoStop}%
\bibitem [{\citenamefont {Price}\ \emph {et~al.}(2025)\citenamefont {Price} \emph {et~al.}}]{price2024gencast}%
  \BibitemOpen
  \bibfield  {author} {\bibinfo {author} {\bibfnamefont {I.}~\bibnamefont {Price}} \emph {et~al.},\ }\bibfield  {title} {\bibinfo {title} {Probabilistic weather forecasting with machine learning},\ }\href {https://doi.org/10.1038/s41586-024-08252-9} {\bibfield  {journal} {\bibinfo  {journal} {Nature}\ }\textbf {\bibinfo {volume} {637}},\ \bibinfo {pages} {84} (\bibinfo {year} {2025})}\BibitemShut {NoStop}%
\bibitem [{\citenamefont {Mochizuki}\ \emph {et~al.}(2018)\citenamefont {Mochizuki} \emph {et~al.}}]{Mochizuki2018snet}%
  \BibitemOpen
  \bibfield  {author} {\bibinfo {author} {\bibfnamefont {M.}~\bibnamefont {Mochizuki}} \emph {et~al.},\ }\bibfield  {title} {\bibinfo {title} {{S-Net} project: performance of a large-scale seafloor observation network for preventing and reducing seismic and tsunami disasters},\ }in\ \href {https://doi.org/10.1109/OCEANSKOBE.2018.8558823} {\emph {\bibinfo {booktitle} {2018 OCEANS - MTS/IEEE Kobe Techno-Oceans (OTO)}}}\ (\bibinfo {year} {2018})\ pp.\ \bibinfo {pages} {1--4}\BibitemShut {NoStop}%
\bibitem [{\citenamefont {Fujii}\ \emph {et~al.}(2011)\citenamefont {Fujii}, \citenamefont {Satake}, \citenamefont {Sakai}, \citenamefont {Shinohara},\ and\ \citenamefont {Kanazawa}}]{fujii2011tohoku}%
  \BibitemOpen
  \bibfield  {author} {\bibinfo {author} {\bibfnamefont {Y.}~\bibnamefont {Fujii}}, \bibinfo {author} {\bibfnamefont {K.}~\bibnamefont {Satake}}, \bibinfo {author} {\bibfnamefont {S.}~\bibnamefont {Sakai}}, \bibinfo {author} {\bibfnamefont {M.}~\bibnamefont {Shinohara}},\ and\ \bibinfo {author} {\bibfnamefont {T.}~\bibnamefont {Kanazawa}},\ }\bibfield  {title} {\bibinfo {title} {Tsunami source of the 2011 off the {Pacific} coast of {Tohoku} earthquake},\ }\href {https://doi.org/10.5047/eps.2011.06.010} {\bibfield  {journal} {\bibinfo  {journal} {Earth Planets Space}\ }\textbf {\bibinfo {volume} {63}},\ \bibinfo {pages} {815} (\bibinfo {year} {2011})}\BibitemShut {NoStop}%
\bibitem [{\citenamefont {Satake}\ \emph {et~al.}(2013)\citenamefont {Satake}, \citenamefont {Fujii}, \citenamefont {Harada},\ and\ \citenamefont {Namegaya}}]{satake2013time}%
  \BibitemOpen
  \bibfield  {author} {\bibinfo {author} {\bibfnamefont {K.}~\bibnamefont {Satake}}, \bibinfo {author} {\bibfnamefont {Y.}~\bibnamefont {Fujii}}, \bibinfo {author} {\bibfnamefont {T.}~\bibnamefont {Harada}},\ and\ \bibinfo {author} {\bibfnamefont {Y.}~\bibnamefont {Namegaya}},\ }\bibfield  {title} {\bibinfo {title} {Time and space distribution of coseismic slip of the 2011 {Tohoku} earthquake as inferred from tsunami waveform data},\ }\href {https://doi.org/10.1785/0120120122} {\bibfield  {journal} {\bibinfo  {journal} {Bull. Seismol. Soc. Am.}\ }\textbf {\bibinfo {volume} {103}},\ \bibinfo {pages} {1473} (\bibinfo {year} {2013})}\BibitemShut {NoStop}%
\bibitem [{\citenamefont {Kawai}\ \emph {et~al.}(2011)\citenamefont {Kawai}, \citenamefont {Satoh}, \citenamefont {Kawaguchi},\ and\ \citenamefont {Seki}}]{kawai2012nowphas}%
  \BibitemOpen
  \bibfield  {author} {\bibinfo {author} {\bibfnamefont {H.}~\bibnamefont {Kawai}}, \bibinfo {author} {\bibfnamefont {M.}~\bibnamefont {Satoh}}, \bibinfo {author} {\bibfnamefont {K.}~\bibnamefont {Kawaguchi}},\ and\ \bibinfo {author} {\bibfnamefont {K.}~\bibnamefont {Seki}},\ }\bibfield  {title} {\bibinfo {title} {The 2011 off the {Pacific} coast of {Tohoku} earthquake tsunami observed by {GPS} buoys},\ }\href {https://doi.org/10.2208/kaigan.67.I\_1291} {\bibfield  {journal} {\bibinfo  {journal} {J. Coast. Eng., JSCE}\ }\textbf {\bibinfo {volume} {67}},\ \bibinfo {pages} {I\_1291} (\bibinfo {year} {2011})}\BibitemShut {NoStop}%
\bibitem [{\citenamefont {Vaswani}\ \emph {et~al.}(2017)\citenamefont {Vaswani} \emph {et~al.}}]{vaswani2017attention}%
  \BibitemOpen
  \bibfield  {author} {\bibinfo {author} {\bibfnamefont {A.}~\bibnamefont {Vaswani}} \emph {et~al.},\ }\bibfield  {title} {\bibinfo {title} {Attention is all you need},\ }in\ \href@noop {} {\emph {\bibinfo {booktitle} {Adv. Neural Inf. Process. Syst.}}},\ Vol.~\bibinfo {volume} {30}\ (\bibinfo {year} {2017})\BibitemShut {NoStop}%
\bibitem [{\citenamefont {Leutbecher}\ and\ \citenamefont {Palmer}(2008)}]{Leutbecher2008}%
  \BibitemOpen
  \bibfield  {author} {\bibinfo {author} {\bibfnamefont {M.}~\bibnamefont {Leutbecher}}\ and\ \bibinfo {author} {\bibfnamefont {T.~N.}\ \bibnamefont {Palmer}},\ }\bibfield  {title} {\bibinfo {title} {Ensemble forecasting},\ }\href {https://doi.org/10.1016/j.jcp.2007.02.014} {\bibfield  {journal} {\bibinfo  {journal} {J. Comput. Phys.}\ }\textbf {\bibinfo {volume} {227}},\ \bibinfo {pages} {3515} (\bibinfo {year} {2008})}\BibitemShut {NoStop}%
\bibitem [{\citenamefont {Baba}\ \emph {et~al.}(2015)\citenamefont {Baba} \emph {et~al.}}]{baba2015parallel}%
  \BibitemOpen
  \bibfield  {author} {\bibinfo {author} {\bibfnamefont {T.}~\bibnamefont {Baba}} \emph {et~al.},\ }\bibfield  {title} {\bibinfo {title} {Parallel implementation of dispersive tsunami wave modeling with a nesting algorithm for the 2011 {Tohoku} tsunami},\ }\href {https://doi.org/10.1007/s00024-015-1049-2} {\bibfield  {journal} {\bibinfo  {journal} {Pure Appl. Geophys.}\ }\textbf {\bibinfo {volume} {172}},\ \bibinfo {pages} {3455} (\bibinfo {year} {2015})}\BibitemShut {NoStop}%
\bibitem [{\citenamefont {Mori}\ \emph {et~al.}(2011)\citenamefont {Mori}, \citenamefont {Takahashi}, \citenamefont {Yasuda},\ and\ \citenamefont {Yanagisawa}}]{mori2011survey}%
  \BibitemOpen
  \bibfield  {author} {\bibinfo {author} {\bibfnamefont {N.}~\bibnamefont {Mori}}, \bibinfo {author} {\bibfnamefont {T.}~\bibnamefont {Takahashi}}, \bibinfo {author} {\bibfnamefont {T.}~\bibnamefont {Yasuda}},\ and\ \bibinfo {author} {\bibfnamefont {H.}~\bibnamefont {Yanagisawa}},\ }\bibfield  {title} {\bibinfo {title} {Survey of 2011 {Tohoku} earthquake tsunami inundation and run-up},\ }\bibfield  {journal} {\bibinfo  {journal} {Geophys. Res. Lett.}\ }\textbf {\bibinfo {volume} {38}},\ \href {https://doi.org/10.1029/2011GL049210} {10.1029/2011GL049210} (\bibinfo {year} {2011})\BibitemShut {NoStop}%
\bibitem [{\citenamefont {{{Geospatial Information Authority of Japan}}}()}]{gsi2011tsunami}%
  \BibitemOpen
  \bibfield  {author} {\bibinfo {author} {\bibnamefont {{{Geospatial Information Authority of Japan}}}},\ }\href {https://maps.gsi.go.jp/development/ichiran.html} {\bibinfo {title} {Tsunami inundation area of the 2011 off the {Pacific} coast of {Tohoku} earthquake}},\ \bibinfo {note} {accessed: 2026-07-23}\BibitemShut {NoStop}%
\bibitem [{\citenamefont {{{National Research Institute for Earth Science and Disaster Resilience (NIED)}}}()}]{bosai_snet_info}%
  \BibitemOpen
  \bibfield  {author} {\bibinfo {author} {\bibnamefont {{{National Research Institute for Earth Science and Disaster Resilience (NIED)}}}},\ }\href {https://www.seafloor.bosai.go.jp/st_info/} {\bibinfo {title} {Seafloor observation network for earthquakes and tsunamis}},\ \bibinfo {note} {accessed: 2026-07-23}\BibitemShut {NoStop}%
\bibitem [{\citenamefont {Tanioka}(2020)}]{tanioka2020improvement}%
  \BibitemOpen
  \bibfield  {author} {\bibinfo {author} {\bibfnamefont {Y.}~\bibnamefont {Tanioka}},\ }\bibfield  {title} {\bibinfo {title} {Improvement of near-field tsunami forecasting method using ocean-bottom pressure sensor network ({S-net})},\ }\href {https://doi.org/10.1186/s40623-020-01268-1} {\bibfield  {journal} {\bibinfo  {journal} {Earth Planets Space}\ }\textbf {\bibinfo {volume} {72}},\ \bibinfo {pages} {132} (\bibinfo {year} {2020})}\BibitemShut {NoStop}%
\bibitem [{\citenamefont {Tsushima}\ \emph {et~al.}(2012)\citenamefont {Tsushima}, \citenamefont {Hino}, \citenamefont {Tanioka}, \citenamefont {Imamura},\ and\ \citenamefont {Fujimoto}}]{tsushima2012tsunami}%
  \BibitemOpen
  \bibfield  {author} {\bibinfo {author} {\bibfnamefont {H.}~\bibnamefont {Tsushima}}, \bibinfo {author} {\bibfnamefont {R.}~\bibnamefont {Hino}}, \bibinfo {author} {\bibfnamefont {Y.}~\bibnamefont {Tanioka}}, \bibinfo {author} {\bibfnamefont {F.}~\bibnamefont {Imamura}},\ and\ \bibinfo {author} {\bibfnamefont {H.}~\bibnamefont {Fujimoto}},\ }\bibfield  {title} {\bibinfo {title} {Tsunami waveform inversion incorporating permanent seafloor deformation and its application to tsunami forecasting},\ }\bibfield  {journal} {\bibinfo  {journal} {J. Geophys. Res. Solid Earth}\ }\textbf {\bibinfo {volume} {117}},\ \href {https://doi.org/10.1029/2011JB008877} {10.1029/2011JB008877} (\bibinfo {year} {2012})\BibitemShut {NoStop}%
\bibitem [{\citenamefont {Suppasri}\ \emph {et~al.}(2013)\citenamefont {Suppasri} \emph {et~al.}}]{suppasri2013building}%
  \BibitemOpen
  \bibfield  {author} {\bibinfo {author} {\bibfnamefont {A.}~\bibnamefont {Suppasri}} \emph {et~al.},\ }\bibfield  {title} {\bibinfo {title} {Building damage characteristics based on surveyed data and fragility curves of the 2011 {Great East Japan} tsunami},\ }\href {https://doi.org/10.1007/s11069-012-0487-8} {\bibfield  {journal} {\bibinfo  {journal} {Nat. Hazards}\ }\textbf {\bibinfo {volume} {66}},\ \bibinfo {pages} {319} (\bibinfo {year} {2013})}\BibitemShut {NoStop}%
\bibitem [{\citenamefont {Sekimoto}\ \emph {et~al.}(2013)\citenamefont {Sekimoto} \emph {et~al.}}]{Sekimoto2013}%
  \BibitemOpen
  \bibfield  {author} {\bibinfo {author} {\bibfnamefont {Y.}~\bibnamefont {Sekimoto}} \emph {et~al.},\ }\bibfield  {title} {\bibinfo {title} {Data mobilization by digital archiving of the {Great East Japan} earthquake survey},\ }\href {https://doi.org/10.5638/thagis.21.87} {\bibfield  {journal} {\bibinfo  {journal} {Theory Appl. GIS}\ }\textbf {\bibinfo {volume} {21}},\ \bibinfo {pages} {87} (\bibinfo {year} {2013})}\BibitemShut {NoStop}%
\bibitem [{\citenamefont {Koshimura}\ \emph {et~al.}(2009)\citenamefont {Koshimura}, \citenamefont {Namegaya},\ and\ \citenamefont {Yanagisawa}}]{koshimura2009tsunami}%
  \BibitemOpen
  \bibfield  {author} {\bibinfo {author} {\bibfnamefont {S.}~\bibnamefont {Koshimura}}, \bibinfo {author} {\bibfnamefont {Y.}~\bibnamefont {Namegaya}},\ and\ \bibinfo {author} {\bibfnamefont {H.}~\bibnamefont {Yanagisawa}},\ }\bibfield  {title} {\bibinfo {title} {Tsunami fragility: A new measure to identify tsunami damage},\ }\href {https://doi.org/10.20965/jdr.2009.p0479} {\bibfield  {journal} {\bibinfo  {journal} {J. Disaster Res.}\ }\textbf {\bibinfo {volume} {4}},\ \bibinfo {pages} {479} (\bibinfo {year} {2009})}\BibitemShut {NoStop}%
\bibitem [{\citenamefont {Richardson}(2000)}]{richardson2000skill}%
  \BibitemOpen
  \bibfield  {author} {\bibinfo {author} {\bibfnamefont {D.~S.}\ \bibnamefont {Richardson}},\ }\bibfield  {title} {\bibinfo {title} {Skill and relative economic value of the {ECMWF} ensemble prediction system},\ }\href {https://doi.org/10.1002/qj.49712656313} {\bibfield  {journal} {\bibinfo  {journal} {Q. J. R. Meteorol. Soc.}\ }\textbf {\bibinfo {volume} {126}},\ \bibinfo {pages} {649} (\bibinfo {year} {2000})}\BibitemShut {NoStop}%
\bibitem [{\citenamefont {Chikasada}(2020)}]{chikasada2020gttm}%
  \BibitemOpen
  \bibfield  {author} {\bibinfo {author} {\bibfnamefont {N.}~\bibnamefont {Chikasada}},\ }\href {https://doi.org/10.17598/NIED.0021} {\bibinfo {title} {Global tsunami terrain model ({GtTM})}} (\bibinfo {year} {2020})\BibitemShut {NoStop}%
\bibitem [{\citenamefont {Ide}\ \emph {et~al.}(2011)\citenamefont {Ide}, \citenamefont {Baltay},\ and\ \citenamefont {Beroza}}]{ide2011shallow}%
  \BibitemOpen
  \bibfield  {author} {\bibinfo {author} {\bibfnamefont {S.}~\bibnamefont {Ide}}, \bibinfo {author} {\bibfnamefont {A.}~\bibnamefont {Baltay}},\ and\ \bibinfo {author} {\bibfnamefont {G.~C.}\ \bibnamefont {Beroza}},\ }\bibfield  {title} {\bibinfo {title} {Shallow dynamic overshoot and energetic deep rupture in the 2011 {Mw} 9.0 {Tohoku-Oki} earthquake},\ }\href {https://doi.org/10.1126/science.1207020} {\bibfield  {journal} {\bibinfo  {journal} {Science}\ }\textbf {\bibinfo {volume} {332}},\ \bibinfo {pages} {1426} (\bibinfo {year} {2011})}\BibitemShut {NoStop}%
\bibitem [{\citenamefont {Kanamori}\ and\ \citenamefont {Anderson}(1975)}]{kanamori1975theoretical}%
  \BibitemOpen
  \bibfield  {author} {\bibinfo {author} {\bibfnamefont {H.}~\bibnamefont {Kanamori}}\ and\ \bibinfo {author} {\bibfnamefont {D.~L.}\ \bibnamefont {Anderson}},\ }\bibfield  {title} {\bibinfo {title} {Theoretical basis of some empirical relations in seismology},\ }\href@noop {} {\bibfield  {journal} {\bibinfo  {journal} {Bull. Seismol. Soc. Am.}\ }\textbf {\bibinfo {volume} {65}},\ \bibinfo {pages} {1073} (\bibinfo {year} {1975})}\BibitemShut {NoStop}%
\bibitem [{\citenamefont {Murotani}\ \emph {et~al.}(2013)\citenamefont {Murotani}, \citenamefont {Satake},\ and\ \citenamefont {Fujii}}]{murotani2013scaling}%
  \BibitemOpen
  \bibfield  {author} {\bibinfo {author} {\bibfnamefont {S.}~\bibnamefont {Murotani}}, \bibinfo {author} {\bibfnamefont {K.}~\bibnamefont {Satake}},\ and\ \bibinfo {author} {\bibfnamefont {Y.}~\bibnamefont {Fujii}},\ }\bibfield  {title} {\bibinfo {title} {Scaling relations of seismic moment, rupture area, average slip, and asperity size for {M\~{} 9} subduction-zone earthquakes},\ }\href {https://doi.org/https://doi.org/10.1002/grl.50976} {\bibfield  {journal} {\bibinfo  {journal} {Geophys. Res. Lett.}\ }\textbf {\bibinfo {volume} {40}},\ \bibinfo {pages} {5070} (\bibinfo {year} {2013})}\BibitemShut {NoStop}%
\bibitem [{\citenamefont {Allmann}\ and\ \citenamefont {Shearer}(2009)}]{allmann2009global}%
  \BibitemOpen
  \bibfield  {author} {\bibinfo {author} {\bibfnamefont {B.~P.}\ \bibnamefont {Allmann}}\ and\ \bibinfo {author} {\bibfnamefont {P.~M.}\ \bibnamefont {Shearer}},\ }\bibfield  {title} {\bibinfo {title} {Global variations of stress drop for moderate to large earthquakes},\ }\bibfield  {journal} {\bibinfo  {journal} {J. Geophys. Res. Solid Earth}\ }\textbf {\bibinfo {volume} {114}},\ \href {https://doi.org/10.1029/2008JB005821} {10.1029/2008JB005821} (\bibinfo {year} {2009})\BibitemShut {NoStop}%
\bibitem [{\citenamefont {Tsushima}\ \emph {et~al.}(2014)\citenamefont {Tsushima}, \citenamefont {Hino}, \citenamefont {Ohta}, \citenamefont {Iinuma},\ and\ \citenamefont {Miura}}]{tsushima2014tfish}%
  \BibitemOpen
  \bibfield  {author} {\bibinfo {author} {\bibfnamefont {H.}~\bibnamefont {Tsushima}}, \bibinfo {author} {\bibfnamefont {R.}~\bibnamefont {Hino}}, \bibinfo {author} {\bibfnamefont {Y.}~\bibnamefont {Ohta}}, \bibinfo {author} {\bibfnamefont {T.}~\bibnamefont {Iinuma}},\ and\ \bibinfo {author} {\bibfnamefont {S.}~\bibnamefont {Miura}},\ }\bibfield  {title} {\bibinfo {title} {{tFISH/RAPiD}: rapid improvement of near-field tsunami forecasting based on offshore tsunami data by incorporating onshore {GNSS} data},\ }\href {https://doi.org/10.1002/2014GL059863} {\bibfield  {journal} {\bibinfo  {journal} {Geophys. Res. Lett.}\ }\textbf {\bibinfo {volume} {41}},\ \bibinfo {pages} {3390} (\bibinfo {year} {2014})}\BibitemShut {NoStop}%
\bibitem [{\citenamefont {Ohta}\ \emph {et~al.}(2012)\citenamefont {Ohta} \emph {et~al.}}]{ohta2012quasi}%
  \BibitemOpen
  \bibfield  {author} {\bibinfo {author} {\bibfnamefont {Y.}~\bibnamefont {Ohta}} \emph {et~al.},\ }\bibfield  {title} {\bibinfo {title} {Quasi real-time fault model estimation for near-field tsunami forecasting based on {RTK-GPS} analysis: application to the 2011 {Tohoku-Oki} earthquake ({Mw} 9.0)},\ }\bibfield  {journal} {\bibinfo  {journal} {J. Geophys. Res. Solid Earth}\ }\textbf {\bibinfo {volume} {117}},\ \href {https://doi.org/10.1029/2011JB008750} {10.1029/2011JB008750} (\bibinfo {year} {2012})\BibitemShut {NoStop}%
\bibitem [{\citenamefont {Song}\ \emph {et~al.}(2021)\citenamefont {Song}, \citenamefont {Meng},\ and\ \citenamefont {Ermon}}]{song2020denoising}%
  \BibitemOpen
  \bibfield  {author} {\bibinfo {author} {\bibfnamefont {J.}~\bibnamefont {Song}}, \bibinfo {author} {\bibfnamefont {C.}~\bibnamefont {Meng}},\ and\ \bibinfo {author} {\bibfnamefont {S.}~\bibnamefont {Ermon}},\ }\bibfield  {title} {\bibinfo {title} {Denoising diffusion implicit models},\ }\bibfield  {booktitle} {\emph {\bibinfo {booktitle} {Int. Conf. Learn. Represent.}},\ }\href {http://dblp.uni-trier.de/db/conf/iclr/iclr2021.html#SongME21} {\  (\bibinfo {year} {2021})}\BibitemShut {NoStop}%
\bibitem [{\citenamefont {Devlin}\ \emph {et~al.}(2019)\citenamefont {Devlin}, \citenamefont {Chang}, \citenamefont {Lee},\ and\ \citenamefont {Toutanova}}]{devlin2019bert}%
  \BibitemOpen
  \bibfield  {author} {\bibinfo {author} {\bibfnamefont {J.}~\bibnamefont {Devlin}}, \bibinfo {author} {\bibfnamefont {M.-W.}\ \bibnamefont {Chang}}, \bibinfo {author} {\bibfnamefont {K.}~\bibnamefont {Lee}},\ and\ \bibinfo {author} {\bibfnamefont {K.}~\bibnamefont {Toutanova}},\ }\bibfield  {title} {\bibinfo {title} {{BERT}: pre-training of deep bidirectional transformers for language understanding},\ }in\ \href {https://doi.org/10.18653/V1/N19-1423} {\emph {\bibinfo {booktitle} {Proc. 2019 Conf. North Am. Chapter Assoc. Comput. Linguist.}}}\ (\bibinfo {year} {2019})\ pp.\ \bibinfo {pages} {4171--4186}\BibitemShut {NoStop}%
\bibitem [{\citenamefont {van~den Oord}\ \emph {et~al.}(2018)\citenamefont {van~den Oord}, \citenamefont {Li},\ and\ \citenamefont {Vinyals}}]{oord2018representation}%
  \BibitemOpen
  \bibfield  {author} {\bibinfo {author} {\bibfnamefont {A.}~\bibnamefont {van~den Oord}}, \bibinfo {author} {\bibfnamefont {Y.}~\bibnamefont {Li}},\ and\ \bibinfo {author} {\bibfnamefont {O.}~\bibnamefont {Vinyals}},\ }\bibfield  {title} {\bibinfo {title} {Representation learning with contrastive predictive coding},\ }\href@noop {} {\bibfield  {journal} {\bibinfo  {journal} {arXiv preprint arXiv:1807.03748}\ } (\bibinfo {year} {2018})}\BibitemShut {NoStop}%
\bibitem [{\citenamefont {Radford}\ \emph {et~al.}(2021)\citenamefont {Radford} \emph {et~al.}}]{radford2021learning}%
  \BibitemOpen
  \bibfield  {author} {\bibinfo {author} {\bibfnamefont {A.}~\bibnamefont {Radford}} \emph {et~al.},\ }\bibfield  {title} {\bibinfo {title} {Learning transferable visual models from natural language supervision},\ }in\ \href {http://proceedings.mlr.press/v139/radford21a.html} {\emph {\bibinfo {booktitle} {Int. Conf. Mach. Learn.}}}\ (\bibinfo {year} {2021})\ pp.\ \bibinfo {pages} {8748--8763}\BibitemShut {NoStop}%
\bibitem [{\citenamefont {Loshchilov}\ and\ \citenamefont {Hutter}(2019)}]{loshchilov2017decoupled}%
  \BibitemOpen
  \bibfield  {author} {\bibinfo {author} {\bibfnamefont {I.}~\bibnamefont {Loshchilov}}\ and\ \bibinfo {author} {\bibfnamefont {F.}~\bibnamefont {Hutter}},\ }\bibfield  {title} {\bibinfo {title} {Decoupled weight decay regularization},\ }\bibfield  {booktitle} {\emph {\bibinfo {booktitle} {Int. Conf. Learn. Represent.}},\ }\href {https://openreview.net/forum?id=Bkg6RiCqY7} {\  (\bibinfo {year} {2019})}\BibitemShut {NoStop}%
\bibitem [{\citenamefont {Ferro}(2014)}]{ferro2014fair}%
  \BibitemOpen
  \bibfield  {author} {\bibinfo {author} {\bibfnamefont {C.~A.~T.}\ \bibnamefont {Ferro}},\ }\bibfield  {title} {\bibinfo {title} {Fair scores for ensemble forecasts},\ }\href {https://doi.org/10.1002/qj.2270} {\bibfield  {journal} {\bibinfo  {journal} {Q. J. R. Meteorol. Soc.}\ }\textbf {\bibinfo {volume} {140}},\ \bibinfo {pages} {1917} (\bibinfo {year} {2014})}\BibitemShut {NoStop}%
\bibitem [{\citenamefont {Zamo}\ and\ \citenamefont {Naveau}(2018)}]{zamo2018estimation}%
  \BibitemOpen
  \bibfield  {author} {\bibinfo {author} {\bibfnamefont {M.}~\bibnamefont {Zamo}}\ and\ \bibinfo {author} {\bibfnamefont {P.}~\bibnamefont {Naveau}},\ }\bibfield  {title} {\bibinfo {title} {Estimation of the continuous ranked probability score with limited information and applications to ensemble weather forecasts},\ }\href {https://doi.org/10.1007/s11004-017-9709-7} {\bibfield  {journal} {\bibinfo  {journal} {Math. Geosci.}\ }\textbf {\bibinfo {volume} {50}},\ \bibinfo {pages} {209} (\bibinfo {year} {2018})}\BibitemShut {NoStop}%
\end{thebibliography}%

\section*{Supplementary Materials}

\setcounter{equation}{0}
\setcounter{figure}{0}
\setcounter{table}{0}
\setcounter{page}{1}

\makeatletter
\renewcommand{\theequation}{S\arabic{equation}}
\renewcommand{\thefigure}{S\arabic{figure}}
\renewcommand{\thetable}{S\arabic{table}}
\renewcommand{\thepage}{S\arabic{page}}
\makeatother

\subsection*{Task definition and probabilistic approach}
The mathematical objective of the tsunami inundation forecasting in this study is to estimate the conditional probability distribution $p_\theta(\mathbf{Y} \mid \mathbf{W}_t)$ of the possible inland inundation depth map $\mathbf{Y}$, conditioned on the time-series waveforms $\mathbf{W}_t$ observed at multiple offshore stations, utilizing a denoising model $r_\theta$. For this purpose, we adopt an ensemble framework that draws multiple independent inundation scenarios $\hat{\mathbf{Y}}_m \sim p_\theta(\mathbf{Y} \mid \mathbf{W}_t)$ for $m=1, 2, \dots, M$ from the estimated probability distribution (generating $M=51$ members in this study).

The input space is $\mathbf{W}_t \in \mathbb{R}^{C \times L}$, where $C$ represents the total number of observation sensors. Here, we use water level waveforms from a total of 150 points---comprising 145 points from the S-net observation network located within the analysis domain \cite{Mochizuki2018snet} and 5 points from NOWPHAS \cite{kawai2012nowphas}---or 5 points from NOWPHAS alone. Regardless of the observation window duration, these waveforms are resampled to a uniform sequence length of $L=256$ using cubic spline interpolation and fed as a $150 \times 256$ or $5 \times 256$ tensor into the waveform encoder, which is custom designed based on the transformer architecture \cite{vaswani2017attention}.
Correspondingly, the output is a tensor $\mathbf{Y} \in \mathbb{R}^{H \times W}$, where $H \times W$ refers to the spatial grid size of $512 \times 512$ pixels discretizing the target area. 

Assuming real-time operational forecasting, the system updates the input $\mathbf{W}_t$ based on the extended observation window as the elapsed time $t$ postearthquake increases (2.5, 5, 10, 15, 20, 25, 30, 35, 40, and 50 minutes in this study), and the data accumulate. It executes inference sequentially with independent models at each time step. Each model dynamically evaluates the potential inundation patterns that are physically consistent with the observations at that moment.

\subsection*{Stochastic tsunami source generation} \label{app:stochastic_source_generation}
For data generation in this study, we adopted the subfault model of Satake et al. (2013) \cite{satake2013time}, which consists of 55 subfaults (an $11 \times 5$ grid) along the Japan Trench (Fig.~\ref{fig:overview}A). To simulate the complex rupture propagation process characteristic of megathrust earthquakes \cite{ide2011shallow}, we modeled the temporal evolution of the slip amount on each subfault in 30-second increments, following the approach of Satake et al. (2013) \cite{satake2013time}.

\begin{figure*}
    \centering
    \includegraphics[page=2, width=\linewidth]{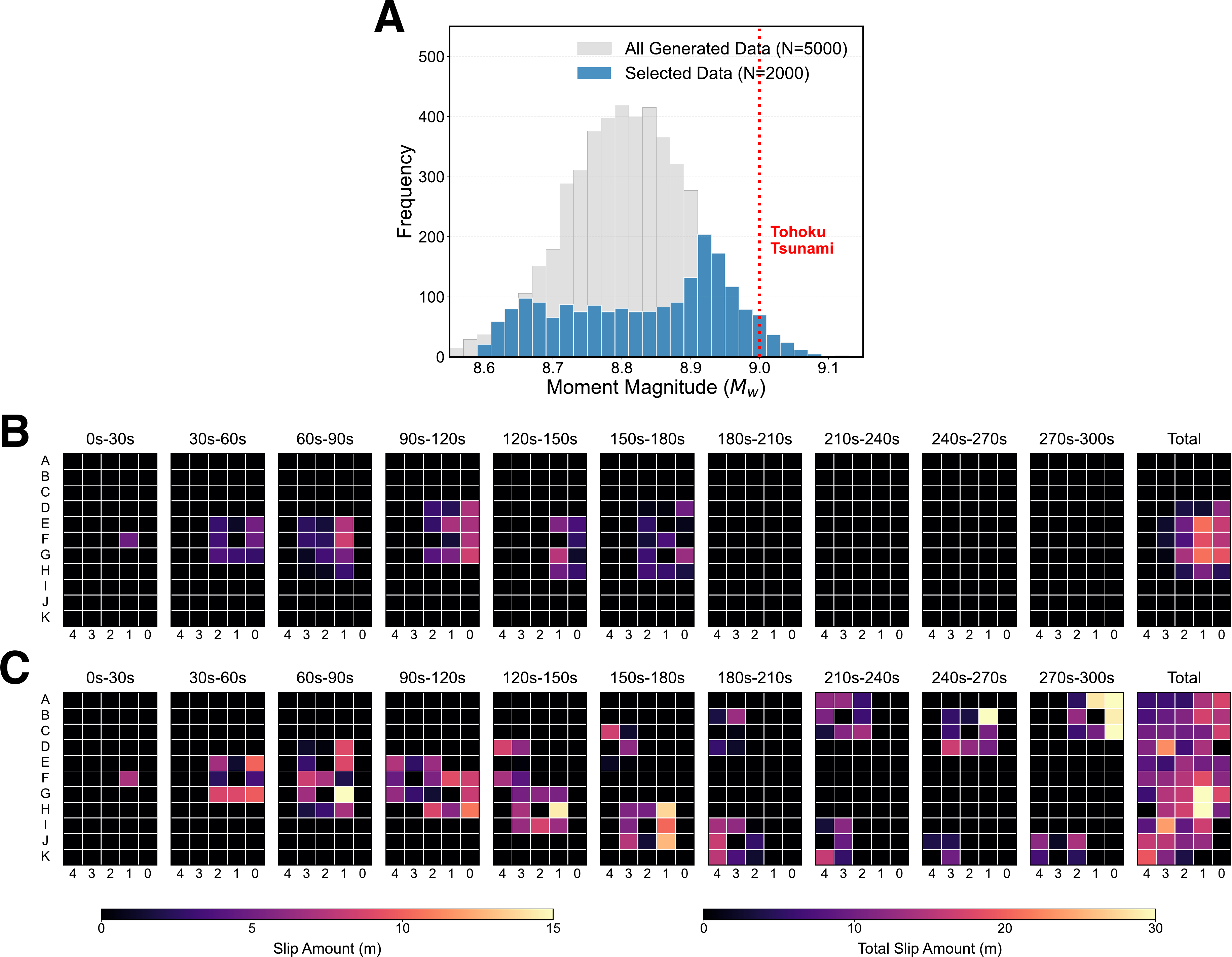}
\caption{\textbf{Tsunami source scenarios generated by the stochastic branching propagation model and the statistical properties of the resulting dataset.} \textbf{A}, Frequency distribution of the moment magnitude ($M_w$) within the generated tsunami source scenarios. The gray area represents all the initially randomly generated data from the stochastic algorithm ($N=5{,}000$), while the blue area represents the final dataset, which was equalized across all the magnitude bands to eliminate training bias ($N=2{,}000$). The red dotted line indicates the magnitude of the 2011 Tohoku-oki earthquake ($M_w$ 9.0). \textbf{B} and \textbf{C}, Examples of rupture propagation processes synthesized by the stochastic algorithm, based on the subfault model structure (11$\times$5 grid, rows A--K, columns 0--4) proposed by Satake et al. \cite{satake2013time}. The actual locations of the subfaults are shown in Fig.~\ref{fig:overview}A. \textbf{B}, Dynamic evolution of the slip distribution in 30-second increments, generated assuming a short rupture duration (minimum of 6 steps, 180-second duration). \textbf{C}, Evolution of a slip distribution with complex spatiotemporal heterogeneity across 10 steps (300-second duration).}

    \label{fig:ed_fault_propagation}
\end{figure*}

First, the initial rupture point is randomly selected from the central region (rows D--H, columns 1--3 in Fig.~\ref{fig:overview}A) of the fault plane and is assigned a random slip amount.
To simulate the spatial rupture propagation of the fault, randomly generated slip amounts are subsequently assigned to the eight adjacent subfaults. Furthermore, among those eight subfaults, the two with the largest slip amounts are selected as the rupture fronts for the next step, with the slip recursively propagating to the surroundings, where each active subfault acts as a new starting point. By repeating this process, we constructed a stochastic branching rupture model that generates heterogeneous and diverse slip distributions on the fault plane. Representative examples of the rupture propagation process generated by this model are shown in Fig.~\ref{fig:ed_fault_propagation}B, C.

The number of 30-second time steps, $N_{step}$, which defines the overall duration of the rupture, was randomly varied for each earthquake scenario. Specifically, this value ranged from a minimum of six steps (180 seconds, including the initial step) to a maximum of ten steps (300 seconds), with the latter matching the dynamic source model of the 2011 Tohoku-oki earthquake proposed by Satake et al. (2013) \cite{satake2013time}.

Furthermore, we applied a base weight to the slip amounts in the shallow, trench-proximal subfaults (columns 0--1), which is twice the weight assigned to the other regions. This application allowed the massive, localized slip patches near the trench axis---characteristic of megathrust earthquakes \cite{satake2013time}---to emerge spontaneously through the cumulative process of stochastic branching propagation.

To ensure the physical plausibility of the final slip distribution, we scaled the overall amplitudes to satisfy the empirical scaling relation $M_0 = \mathcal{C} \cdot S^{3/2}$ between the seismic moment $M_0$ and the ruptured fault area $S$. This scaling is based on the theoretical circular crack model \cite{kanamori1975theoretical} and has been empirically validated for megathrust earthquakes in subduction zones \cite{murotani2013scaling}. Here, the fault area $S$ was defined as the total area of subfaults that exhibited nonzero final slip. To account for the natural interevent variability \cite{allmann2009global}, the stress drop $\Delta\sigma$ was randomly sampled from the range of $[3.0, 5.0]\,\mathrm{MPa}$, yielding the scaling coefficient $\mathcal{C} = \frac{16 \Delta\sigma}{7 \pi^{3/2}}$ used to scale the overall slip amplitudes.

To mitigate bias in the generated scenarios, we employed a balanced sampling strategy for the generated events based on the moment magnitude, $M_w = (\log_{10} M_0 - 9.1) / 1.5$. Specifically, we divided $M_w$ into bins of a fixed width and aimed to extract an equal number of samples from each bin. However, owing to an insufficient number of samples at the tails of the distribution (representing exceptionally large or small events), we compensated for the shortfall by drawing additional samples from adjacent bins closer to the mean $M_w$, where data were more abundant. Through this procedure, we constructed a dataset totalling 2,000 cases with a data density as uniform as possible across all magnitude bands while fully utilizing the rare, massive events (the $M_w$ distribution of the constructed dataset is shown in Fig.~\ref{fig:ed_fault_propagation}A).

Next, we executed tsunami inundation simulations for all the generated earthquake scenarios. We retained 1,993 cases for the final dataset, excluding seven cases where numerical calculations became unstable because of the use of a uniform fixed time step. The retained cases were subsequently partitioned into 1,613 training cases, 180 validation cases, and 200 test cases. Statistical analysis of the training data confirmed that the 2011 Tohoku-oki earthquake represents an extreme event positioned significantly above the median (Fig.~\ref{fig:ed_dataset_stats}A, D). Furthermore, the generated dataset covers a wide range of cases, from small events with almost no inundation to large events with inundation depths exceeding 10 m (Fig.~\ref{fig:ed_dataset_stats}E, F).

\subsection*{Data generation, preprocessing, and postprocessing}
For data generation, we employed the nonlinear longwave equation solver JAGURS \cite{baba2015parallel} to reproduce the tsunami generation, propagation, and run-up processes using a five-level nested grid system (15\,m, 45\,m, 135\,m, 405\,m, and 1215\,m). We generated offshore time-series waveforms $\mathbf{W}_t$ simulated with a time step of $\Delta t = 0.5$\,s, alongside onshore target maps $\mathbf{Y}$ representing the maximum inundation depth or tsunami arrival time over a four-hour simulation period.

When the target map $\mathbf{Y}$ represents the maximum inundation depth, to focus the learning strictly on onshore tsunami inundation, we first applied a masking process that assigned a constant background value of $-1.0\,\mathrm{m}$ to the ocean domain, identical to the treatment of noninundated land areas. Furthermore, to accentuate the boundary between inundated and noninundated regions, we uniformly added a $2.0\,\mathrm{m}$ offset to all inundated pixels. This addition induced an artificial $3.0\,\mathrm{m}$ discontinuity (the step from $-1.0\,\mathrm{m}$ to $2.0\,\mathrm{m}$) at the inundation front, enhancing the model's ability to delineate the inundation boundary.
Referencing the maximum observed values in the training dataset, the inundation depth was subsequently clipped at an upper limit of $14.0\,\mathrm{m}$ ($16.0\,\mathrm{m}$ including the offset). To format these physical quantities as image representations for training, the value range $[-1.0\,\mathrm{m}, 16.0\,\mathrm{m}]$ was linearly scaled to the range of $[0.25, 0.75]$. Preliminary experiments demonstrated that this specific scaling range improves the model's prediction accuracy compared with that of a standard mapping to $[0.0, 1.0]$.

Data generation was also performed for cases where the target map $\mathbf{Y}$ represents the tsunami arrival time. In these instances, noninundated and ocean areas were assigned a constant background value of $16,000\,\mathrm{s}$, and the physical quantity range $[0.0\,\mathrm{s}, 16,000\,\mathrm{s}]$ was linearly scaled to the range of $[0.25, 0.75]$. The four-hour ($14,400\,\mathrm{s}$) simulation duration inherently provides a gap of at least $1,600\,\mathrm{s}$ from the background value ($16,000\,\mathrm{s}$) at the inundation front, eliminating the need for an artificial offset.

During inference, the predicted image representations were inversely scaled back to physical quantities and postprocessed to ensure physical validity. Specifically, maximum inundation depth predictions were clipped at a lower bound of $0.0\,\mathrm{m}$ to eliminate nonphysical negative values. Similarly, tsunami arrival time predictions were capped at the simulation limit of $14,400\,\mathrm{s}$, yielding the final prediction results.

\subsubsection*{Crustal deformation correction for ocean bottom pressure gauge data}
The water depth variations $h_f(x;t)$ recorded by ocean bottom pressure gauges (OBPGs), such as S-net, include not only pure tsunami-induced sea surface displacements $\eta(x;t)$ but also the influence of the vertical movement of the sensors themselves due to coseismic seafloor deformation $B(x;t)$ \cite{tsushima2012tsunami}. This relationship is expressed as $h_f(x;t) = \eta(x;t) - B(x;t)$. During the data generation process, we extracted the vertical displacement at each observation point from the initial crustal deformation and subtracted it from the simulated tsunami waveform, thereby emulating the characteristic baseline shifts observed by actual OBPGs.

\subsubsection*{Baseline correction for GPS wave buoy data}
Since GPS wave buoys such as NOWPHAS installed in coastal areas measure sea surface variations as a relative vertical position with respect to a nearby onshore base station, correcting for apparent baseline shifts caused by the subsidence or uplift of the onshore base station itself is necessary. The extent of this crustal deformation at the onshore base station can be estimated in real time through the rapid analysis of GNSS observations \cite{tsushima2014tfish, ohta2012quasi}. Therefore, in this study, we did not introduce this effect into the training and test data. In contrast, in the validation using actual observation data from the 2011 Tohoku-oki earthquake, we applied corrections using the actual baseline shift $\Delta h$ observed by NOWPHAS \cite{kawai2012nowphas}. Assuming a fault rupture duration of $T_{\mathrm{rupture}} = 300\,\mathrm{s}$, we incorporated a dynamic correction where the shift increases linearly from $0$ to $\Delta h$ during this period.

\subsection*{Generative framework: latent diffusion for tsunami prediction}
In this study, the tsunami prediction map generation process utilizes the framework of latent diffusion models (LDMs) \cite{rombach2022high}, widely used for image generation. This system consists primarily of the following three standard components and a uniquely designed waveform encoder, which is detailed in the next section.

\paragraph*{Variational autoencoder (VAE)}
Instead of directly processing the target physical quantities (maximum inundation depth and arrival time) in the high-dimensional pixel space, these variables are compressed into a lower-dimensional $64 \times 64$ latent space.
In this system, the pretrained VAE accompanying Stable Diffusion v1-5 \cite{rombach2022high} was utilized for compression without modification. Specifically, the scaled physical quantities were treated as image representations and input as RGB images by duplicating the single-channel values across all three channels. The VAE compresses these $512 \times 512 \times 3$ images into 4-channel latent features ($64 \times 64 \times 4$), reducing the computational cost during training and inference. During inference, the generated latent space features are decoded by the VAE, and the average value across the three output RGB channels is calculated to convert them back into the final physical quantities at a $512 \times 512$ resolution.

\paragraph*{Conditional denoising model (U-Net)} 
We employed the U-Net architecture from Stable Diffusion v1-5 \cite{rombach2022high} to drive the denoising process in the latent space. Initialized with the pretrained weights from the original image generation task, the model was fine-tuned on our tsunami simulation dataset to adapt it for generating tsunami inundation maps. The cross-attention layers within U-Net integrate the output from our custom-designed waveform encoder (detailed below) to condition the denoising process.

\paragraph*{Sampling scheduler (DDIM)}
To accelerate inference, we employed the denoising diffusion implicit model (DDIM) scheduler \cite{song2020denoising}. While training assumes 1,000 diffusion steps, we reduced the number of sampling steps during inference to 50. This adjustment significantly accelerates the generation of ensembles, rendering real-time forecasting feasible. For the stochasticity control parameter $\eta$, we set $\eta = 1$ for both \textsf{GenTEW} and \textsf{GenTEW-Diffusion} to capture the inherent stochasticity of the generative process. Conversely, for \textsf{GenTEW-Perturbed}---which relies entirely on input perturbations to derive its probabilistic forecasts---we set $\eta = 0$ to enforce deterministic sampling.

\subsection*{Conditioning module: transformer-based waveform encoder}
To convert the time-series waveform data into conditioning features for the diffusion model (U-Net), we introduced a custom-designed waveform encoder based on the transformer architecture \cite{vaswani2017attention}. Given a 1-dimensional waveform input with a sequence length of $L=256$ and $C$ channels, a linear layer first projects the $L \times C$ input tensor into a hidden state sequence of shape $L \times d_{\mathrm{model}}$ ($d_{\mathrm{model}}=768$). To retain temporal sequence information, learnable positional embeddings \cite{devlin2019bert} are added to the feature vectors.

The data are subsequently passed through a stack of four transformer encoder layers. Each layer consists of a multihead self-attention mechanism with 12 attention heads and a feed-forward network (FFN) with an internal dimension of $d_{\mathrm{ff}}=3072$. Layer normalization---along with dropout (at a rate of 0.1) and residual connections---is applied to each sublayer, with ReLU serving as the activation function in the FFN.

Similar to a sequence of word tokens in a language model, the waveform encoder preserves the original sequence length $L = 256$ and outputs a tensor of shape $256 \times 768$. This design enables the cross-attention layers of U-Net to effectively attend to critical temporal features within the waveform data that have a major impact on inundation---such as the wavelength of the initial wave---during the generation of inundation maps.

\subsection*{Two-stage training: contrastive learning and diffusion fine-tuning}
The training process of our model consists of two stages: (1) contrastive learning of the waveform encoder, which associates waveform signals with spatial inundation patterns, and (2) fine-tuning of the diffusion model, which generates inundation maps conditioned on the features extracted by the waveform encoder.

For the first stage of contrastive representation learning, an image encoder is introduced exclusively during training to extract spatial features from the corresponding inundation map $\mathbf{Y}$. We constructed a custom 7-layer CNN as this image encoder. Each layer consists of a convolutional layer (kernel size 4, stride 2, padding 1) followed by ReLU activation. Through seven successive applications of this layer structure, the spatial resolution is progressively downsampled ($512 \times 512 \rightarrow 256 \times 256 \rightarrow 128 \times 128 \rightarrow \cdots \rightarrow 4 \times 4$), while the number of channels is systematically expanded ($16 \rightarrow 32 \rightarrow 64 \rightarrow 128 \rightarrow 256 \rightarrow 512 \rightarrow 512$) to extract hierarchical spatial features.

The resulting $512 \times 4 \times 4$ feature map is flattened into an 8,192-dimensional vector and projected into a latent vector of dimension $d_{\mathrm{model}}=768$ via a linear layer. 
This projection aligns the spatial representation with the 768-dimensional waveform feature vector, which is obtained by applying temporal mean pooling to the output tensor of the waveform encoder. Treating unmatched waveform-map pairs within the same mini-batch as negative examples, the encoders are jointly optimized by minimizing the symmetric cross-entropy loss (InfoNCE loss, with a temperature parameter $\tau = 0.07$) based on cosine similarity \cite{oord2018representation, radford2021learning}.

In the second stage, U-Net is fine-tuned, while the waveform encoder pretrained in the first stage is kept frozen to serve as a conditioning module. We minimize the following mean squared error (MSE) between the injected standard Gaussian noise, $\boldsymbol{\epsilon} \sim \mathcal{N}(\mathbf{0}, \mathbf{I})$, and the noise predicted by the model:
\begin{equation} 
\mathcal{L} = \mathbb{E} \left[ \| \boldsymbol{\epsilon} - \boldsymbol{\epsilon}_\theta(\mathbf{z}_k, k, \mathbf{Z}_{\mathrm{wave}}) \|^2 \right].
\end{equation}
Here, $k \in \{1, \dots, 1000\}$ is the diffusion step randomly sampled from a uniform distribution during training, $\mathbf{z}_k$ is the noisy latent representation, $\boldsymbol{\epsilon}_\theta$ denotes U-Net, and $\mathbf{Z}_{\mathrm{wave}}$ is the feature tensor output by the waveform encoder.

For optimization in both the first and second stages, we employed the AdamW optimizer \cite{loshchilov2017decoupled}. To manage learning rate scheduling, we applied a decay strategy triggered by validation loss plateaus: upon stagnation, training was paused and resumed with a reduced learning rate. Specifically, optimization progressed through three learning rate stages ($1 \times 10^{-4}, 1 \times 10^{-5}, 1 \times 10^{-6}$) for waveform encoder training and two stages ($1 \times 10^{-4}, 1 \times 10^{-5}$) for U-Net training.

\subsection*{Waveform perturbation and baseline models}
\subsubsection*{Perturbation strategy}
To comprehensively evaluate the uncertainty inherent in tsunami forecasting, we introduced perturbations to the input waveforms inspired by the concept of meteorological ensemble forecasting \cite{Leutbecher2008}. The addition of multiplicative noise to the input waveforms facilitates a calibratable representation of the model's epistemic uncertainty. Concurrently, injecting additive noise accounts for the aleatoric uncertainty caused by observation errors. The perturbed input waveform $\mathbf{W}'_t$ is defined by the following equation:
\begin{equation} 
\mathbf{W}'_t = \mathbf{W}_t + \gamma_{\mathrm{mult}} \cdot |\mathbf{W}_t| \cdot \boldsymbol{\delta}_{\mathrm{long}}(t) + \gamma_{\mathrm{add}} \cdot \boldsymbol{\delta}_{\mathrm{short}}(t).
\end{equation}
The multiplicative component $\boldsymbol{\delta}_{\mathrm{long}}(t)$ is generated by applying a Gaussian filter with a standard deviation $\sigma_{\mathrm{long}}$ to smooth the white noise independently generated for each sensor and is standardized to zero mean and unit variance. To represent the uncertainty of long-term tsunami phenomena, $\sigma_{\mathrm{long}}$ is randomly sampled for each ensemble member between $60\,\mathrm{s}$ and $3600\,\mathrm{s}$, and a common value is applied to all the sensors within the same member. The intensity coefficient $\gamma_{\mathrm{mult}}$ is sampled uniformly from $[0, \alpha]$ for each member, where the hyperparameter $\alpha$ controls the upper limit of the noise intensity; this coefficient is also shared among all the sensors.

\begin{figure*}
    \centering
    \includegraphics[page=4, width=\linewidth]{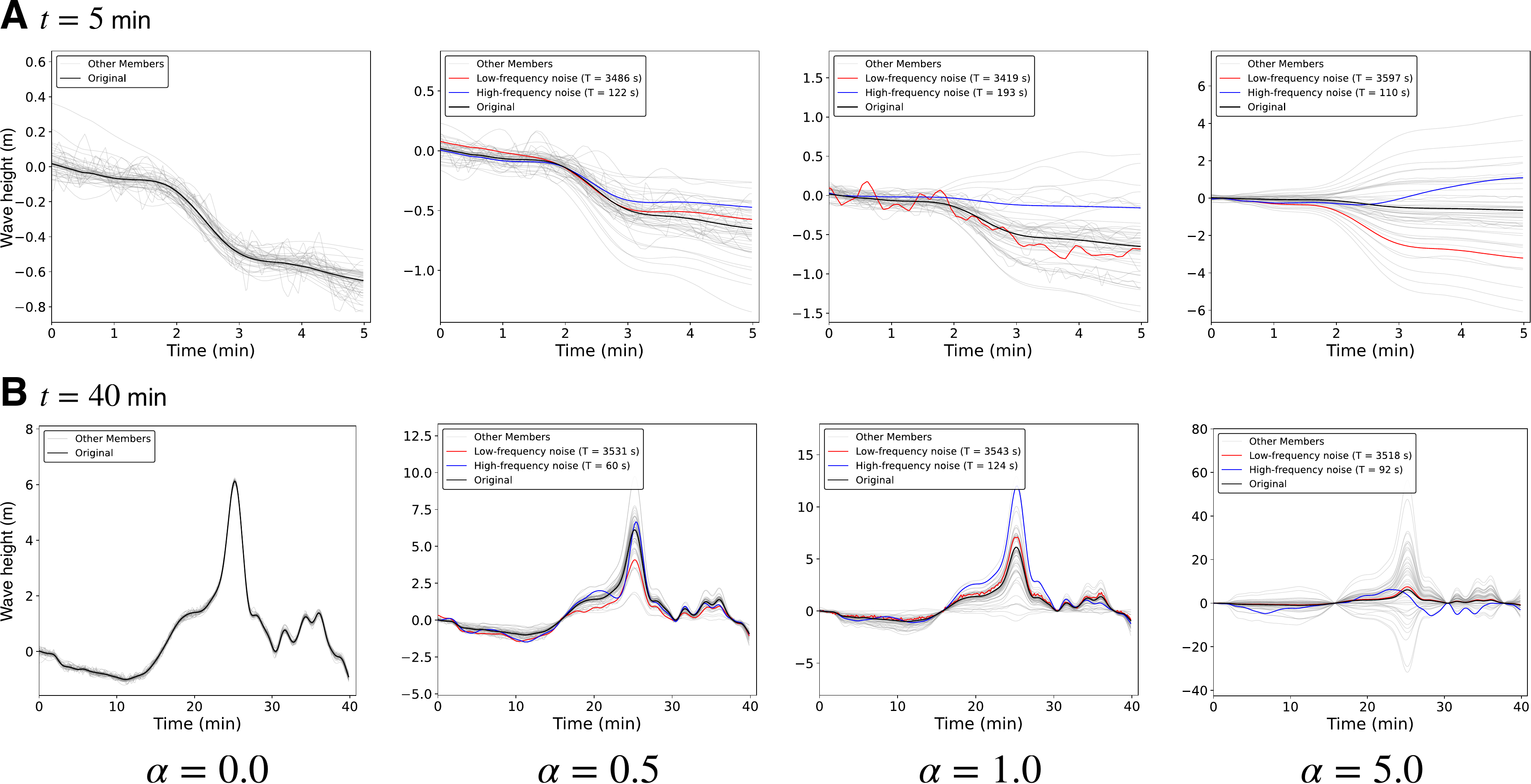}

\caption{\textbf{Input waveforms perturbed with the intensity upper limit parameter $\alpha$.} Examples of perturbed waveforms for ensemble members at 5 and 40 minutes post-event (\textbf{A} and \textbf{B}, respectively) generated by varying the hyperparameter $\alpha$ (0.0, 0.5, 1.0, and 5.0), which controls the upper limit of the multiplicative noise intensity applied to the original waveforms (Original). Short-period additive noise, representing aleatoric uncertainty such as observation errors, is consistently applied even when $\alpha=0.0$. Long-period multiplicative noise, representing the model's epistemic uncertainty, is superimposed onto this additive noise. In the figure, among all members, those with the lowest frequency (longest period) and highest frequency (shortest period) of multiplicative noise are indicated by red and blue lines, respectively, for reference. At $\alpha=1.0$, waveforms with reversed positive and negative signs are included, significantly altering the original waveforms. At $\alpha=5.0$, waveforms reaching nearly 10 times the original wave height are generated, largely obscuring the original waveform information.}

    \label{fig:ed_waveform_perturbation}
\end{figure*}

Moreover, the additive component $\boldsymbol{\delta}_{\mathrm{short}}(t)$ is generated using a Gaussian filter with $\sigma_{\mathrm{short}}$ sampled between $0\,\mathrm{s}$ and $60\,\mathrm{s}$, following the same standardization. Considering instrumental noise, the intensity coefficient $\gamma_{\mathrm{add}}$ is sampled uniformly from a range whose upper limit corresponds to a maximum standard deviation of $0.1\,\mathrm{m}$ in physical space. This setting is grounded in previous research \cite{mulia2022machine} considering the noise characteristics of S-net observation data. Examples of the waveforms with these injected perturbations are shown in Fig.~\ref{fig:ed_waveform_perturbation}.

\subsubsection*{Baseline models}
To evaluate \textsf{GenTEW}, we introduce two baseline models. The first baseline, \textsf{GenTEW-Diffusion}, relies solely on the probabilistic generative capabilities inherent in LDMs \cite{price2024gencast}. For this model, only the additive noise simulating observation errors is added to the input waveforms. During inference, the DDIM stochasticity parameter is set to $\eta = 1$, and an ensemble is generated by varying the seed value of the initial latent noise.

The second baseline, \textsf{GenTEW-Perturbed}, draws inspiration from previous research on diffusion models in weather forecasting \cite{price2024gencast} and isolates and evaluates the impact of input waveform uncertainty on the predictions. For this model, the stochasticity parameter during inference is set to $\eta = 0$, and the initial latent noise is fixed to the same seed value for all members. Consequently, the inference process becomes completely deterministic, ensuring that the ensemble is generated solely by the perturbations applied to the input waveforms.

Note that both baseline models share identical weights for the waveform encoder and U-Net across different observation conditions.

\subsubsection*{Noise intensity optimization}
For \textsf{GenTEW-Perturbed} and \textsf{GenTEW}, both of which involve perturbations to the input waveforms, we optimized the upper limit of the multiplicative noise intensity $\alpha \in [0, 5.0]$ via a grid search. This optimization was performed using an independent validation subset (50 samples) to calibrate the spread-skill ratio toward the ideal value of 1.0. To prevent accuracy degradation due to excessive noise, we identified and adopted the minimum $\alpha$ value that achieved a spread-skill ratio of $\ge 1.0$ at a precision of 0.01.

\begin{figure*}
    \centering
    \includegraphics[page=7, width=1\linewidth]{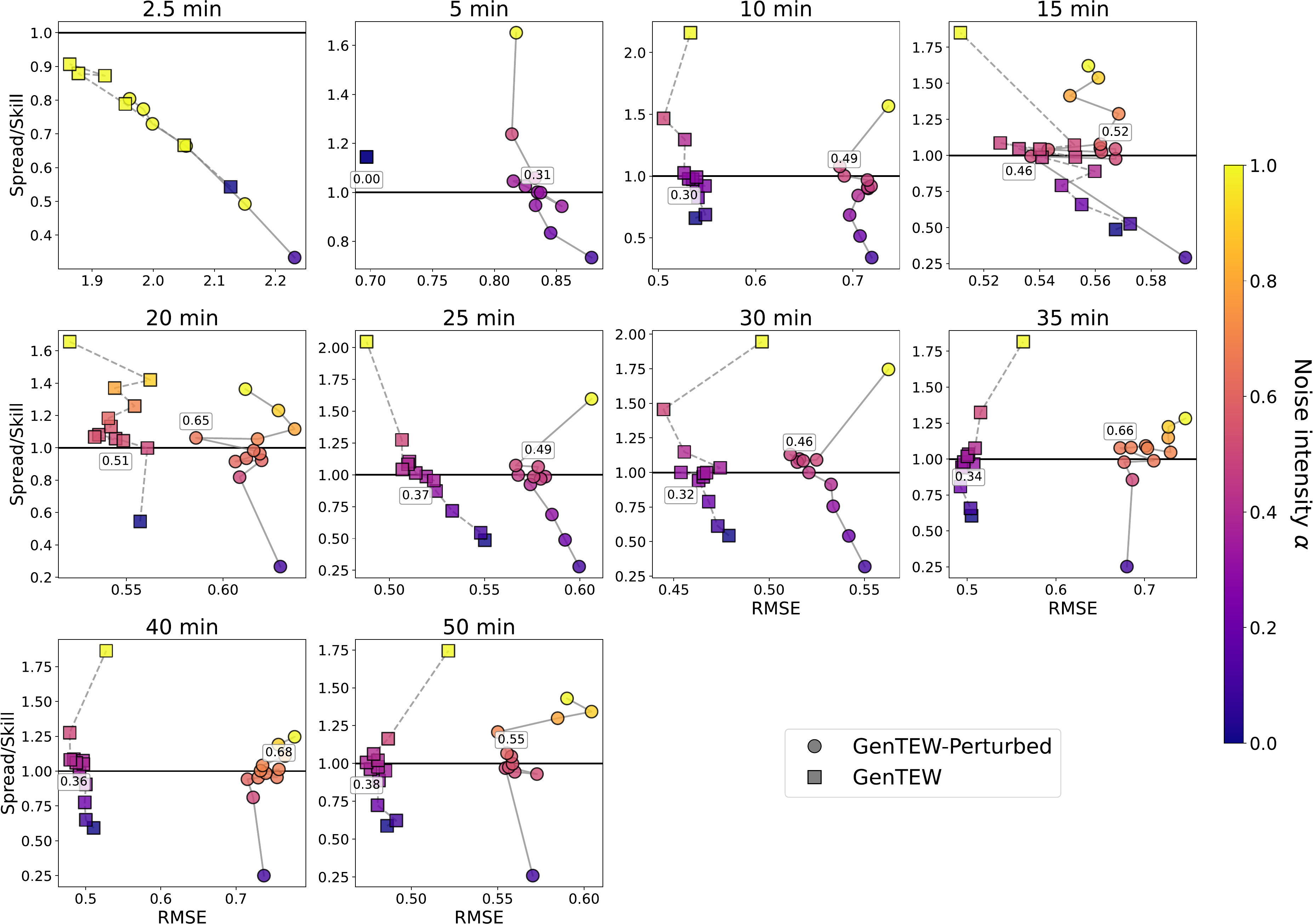}
\caption{\textbf{Optimization of the multiplicative noise intensity upper limit ($\alpha$) and the calibration process via grid search.} Grid search for the multiplicative noise intensity upper limit parameter $\alpha \in [0, 5.0]$ at various observation times (2.5 to 50 minutes post-event) using the validation dataset. The vertical axis represents the spread-skill ratio, the horizontal axis represents RMSE, and the marker color indicates the applied noise intensity level. Circles (solid lines) and squares (dashed lines) represent \textsf{GenTEW-Perturbed} and \textsf{GenTEW}, respectively. The optimal value for $\alpha$ was identified as the minimum noise level where the spread-skill ratio exceeded the ideal value of 1.0, using a search interval of 0.01. In the figure, the corresponding $\alpha$ values are explicitly labeled only for the data points that met this condition and were adopted as the optimal values for each model. However, at early observation stages (2.5 minutes post-event), the spread-skill ratio failed to reach 1.0 even at the maximum search limit ($\alpha=5.0$); therefore, the maximum value of 5.0 was selected. Throughout this search process, \textsf{GenTEW} exhibits a lower RMSE than \textsf{GenTEW-Perturbed} in almost all cases. This finding demonstrates the superior prediction accuracy of \textsf{GenTEW}.}

    \label{fig:ed_noise_optimization}
\end{figure*}

For cases where the spread-skill ratio was $\ge 1.0$ at $\alpha=0.0$, we adopted $\alpha=0.0$. Conversely, particularly in scenarios with short observation windows, there were instances where the spread-skill ratio did not reach 1.0 even when the noise intensity was raised to the upper search limit of $\alpha=5.0$; in such cases, $\alpha=5.0$ was adopted. Since \textsf{GenTEW-Diffusion} relies exclusively on the probabilistic properties inherent in diffusion models, it is not capable of this explicit noise tuning. The results of the grid search are shown in Fig.~\ref{fig:ed_noise_optimization}, and the optimized $\alpha$ values are summarized in Table~\ref{tab:optimization_results}.
Notably, the optimized $\alpha$ values for \textsf{GenTEW} were generally lower than those for \textsf{GenTEW-Perturbed}. This finding demonstrates that \textsf{GenTEW} achieves the target spread-skill ratio with less input perturbation, thereby avoiding excessive degradation of physical waveform information and facilitating superior balance between predictive calibration and accuracy.

\subsection*{Unbiased evaluation metrics}
\subsubsection*{Safety-first estimation for RMSE}
When the RMSE for the maximum inundation depth was calculated, the aggregation was limited to areas where the true inundation depth $D_{\mathrm{true}}$ was positive ($D_{\mathrm{true}} > 0$). Similarly, in the evaluation of the tsunami arrival time, the RMSE aggregation was limited to regions where the tsunami was captured in the ground truth data.
By restricting evaluation to ground-truth hazard areas, we exclude false positives (false alarms) from the error calculation while directly penalizing false negatives (misses). This approach constitutes a safety-first scoring that prioritizes the avoidance of misses, which are critical in disaster prevention systems.

\subsubsection*{Fair estimators for CRPS}
In this study, we adopted the fair estimator $\mathrm{CRPS}_{\mathrm{fair}}$ \cite{ferro2014fair, zamo2018estimation} to calculate the CRPS. This metric is defined by the following equation, using the number of evaluation samples $K$ and the number of valid pixels $|G|_k$ in each sample $k$:
\begin{equation} 
\begin{split}
\mathrm{CRPS}_{\mathrm{fair}} &:= \frac{1}{\sum_{k=1}^K |G|_k} \sum_{k=1}^K \sum_{i=1}^{|G|_k} \Bigg( \frac{1}{M} \sum_m |x_{i,k}^m - y_{i,k}| \\
&\qquad - \frac{1}{2M(M-1)} \sum_{m,m'} |x_{i,k}^m - x_{i,k}^{m'}| \Bigg).
\end{split}
\end{equation}
Here, $k$ is the sample index, $i$ is the valid pixel index, $m$ and $m'$ are the ensemble member indices, and $M$ represents the total number of members. The variables $x_{i,k}^m$ and $y_{i,k}$ denote the predicted value and the corresponding ground truth, respectively. In alignment with the aforementioned safety-first evaluation criteria, the number of valid pixels $|G|_k$ counts only the pixels that were inundated in the ground truth data.

Furthermore, the fair estimator of the RMSE ($\mathrm{RMSE}_{\mathrm{fair}}$) and the spread-skill ratio ($\mathrm{SSR}$) are defined by the following equations \cite{price2024gencast}:
\begin{equation} 
\mathrm{RMSE}_{\mathrm{fair}} := \sqrt{ \frac{1}{\sum_{k=1}^K |G|_k} \sum_{k=1}^K \sum_{i=1}^{|G|_k} (\bar{x}_{i,k} - y_{i,k})^2 - \frac{\bar{s}^2}{M} },
\end{equation}
\begin{equation} 
\mathrm{SSR} := \frac{\bar{s}}{\mathrm{RMSE}_{\mathrm{fair}}}.
\end{equation}
Here, $\bar{x}_{i,k}$ is the ensemble mean value, and $\bar{s}^2$ is the mean ensemble variance:
\begin{equation} 
\bar{s}^2 := \frac{1}{\sum_{k=1}^K |G|_k} \sum_{k=1}^K \sum_{i=1}^{|G|_k} \left[ \frac{1}{M-1} \sum_{m=1}^M (x_{i,k}^m - \bar{x}_{i,k})^2 \right].
\end{equation}
Throughout this paper, all the references to the CRPS, RMSE, and spread-skill ratio denote the above-defined fair estimators.

\subsection*{Practical utility evaluation of ensemble forecasts}
\subsubsection*{Tsunami fragility assessment and hazard risk probability}
To demonstrate the practical utility of the generated ensemble forecasts, we assessed the expected damage probability $P_{\mathrm{risk}}$. To evaluate building vulnerability, we adopted the fragility curve formulation and its corresponding empirical parameters derived by Suppasri et al. \cite{suppasri2013building} based on the 2011 Tohoku-oki earthquake data. In this approach, the cumulative damage probability (fragility curve) of a building, $P_f(D)$, is modeled as the cumulative distribution function (CDF) of a log-normal distribution with respect to the tsunami water depth $D$:
\begin{equation} 
P_f(D) = \Phi \left( \frac{\ln D - \mu}{\sigma} \right).
\end{equation}
Here, $\Phi(\cdot)$ denotes the CDF of the standard normal distribution. The parameters $\mu$ and $\sigma$ represent the mean and standard deviation of the logarithmic water depth, respectively.

To integrate our ensemble forecast uncertainty with this vulnerability model, we formulate the expected damage risk, $P_{\mathrm{risk}}$, by calculating the product of the tsunami exceedance probability $P_{\mathrm{exceed}}(D)$ and the damage probability $P_f(D)$. 
We evaluate this product at 0.1-m intervals and identify the maximum value as follows:
\begin{equation} 
P_{\mathrm{risk}} = \max_D \left[ P_{\mathrm{exceed}}(D) \times P_f(D) \right].
\end{equation}

Moreover, the same framework can be extended to assess human safety risk by incorporating fragility curves for fatality \cite{koshimura2009tsunami}.

\subsubsection*{REV based on the cost-loss decision model}
\begin{figure*}
    \centering
    \includegraphics[page=3, width=\linewidth]{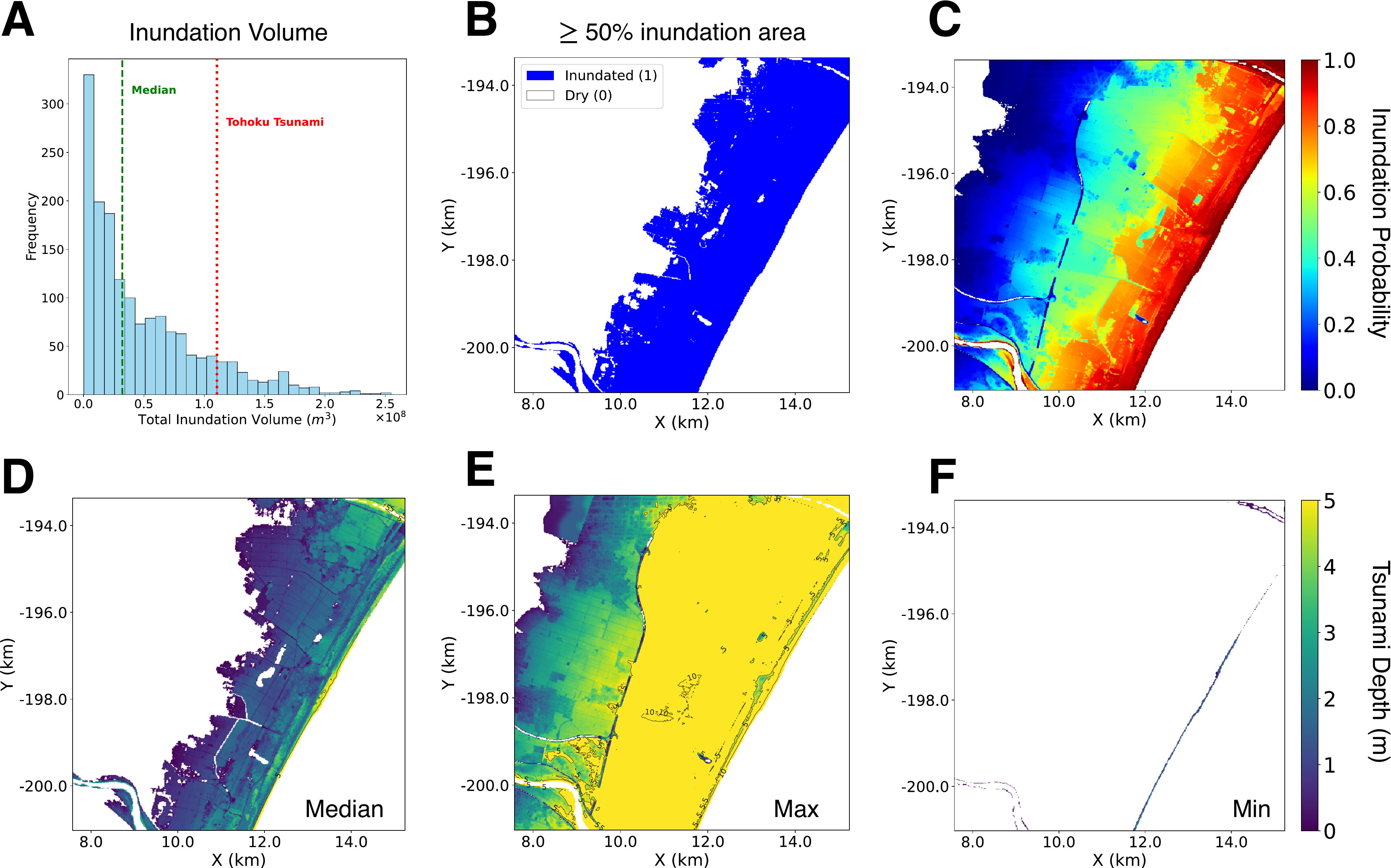}

\caption{\textbf{Spatial and statistical characteristics of the training dataset.} \textbf{A}, Frequency distribution of the total inundation volume (defined as the sum of the product of the maximum inundation depth and pixel area) for the entire training dataset. The green dashed line indicates the median, and the red dotted line indicates the scale corresponding to the 2011 Tohoku-oki earthquake \cite{satake2013time}. \textbf{B}, Areas inundated in at least 50\% of the scenarios across all training data. \textbf{C}, Spatial distribution of the pixelwise inundation probability (the proportion of scenarios in which a given pixel is inundated within the entire dataset). \textbf{D}--\textbf{F}, Inundation depth maps for the scenarios representing the median (\textbf{D}), maximum (\textbf{E}), and minimum (\textbf{F}) when the data are ordered by total inundation volume.}

    \label{fig:ed_dataset_stats}
\end{figure*}

To quantify the economic value of the tsunami inundation forecasts, we introduced the relative economic value (REV) based on a cost-loss decision model \cite{richardson2000skill, price2024gencast}. Here, $\text{Cost}$ represents the cost of taking preventive actions, and $\text{Loss}$ denotes the potential loss incurred if no action is taken. As a baseline for comparison, we employed a static inundation map derived from the average inundation probability across the entire training dataset (Fig.~\ref{fig:ed_dataset_stats}C). For both this baseline and our probabilistic inundation forecasts, we binarized the maps by classifying pixels with an inundation probability of $\ge 50\%$ as inundated (Fig.~\ref{fig:ed_dataset_stats}B). These binarized maps were then compared against the ground truth data to compute the proportions of correct and incorrect predictions. Let $o$ be the true occurrence probability of inundation; the expected expenses for each forecasting scenario are formulated as follows:
\begin{equation} E_{\mathrm{base}} = \mathrm{TP}_{\mathrm{base}} \cdot \text{Cost} + \mathrm{FP}_{\mathrm{base}} \cdot \text{Cost} + \mathrm{FN}_{\mathrm{base}} \cdot \text{Loss},
\end{equation} \begin{equation} E_{\mathrm{forecast}} = \mathrm{TP} \cdot \text{Cost} + \mathrm{FP} \cdot \text{Cost} + \mathrm{FN} \cdot \text{Loss},
\end{equation} \begin{equation} E_{\mathrm{perfect}} = o \cdot \text{Cost}. \end{equation}

Here, $\mathrm{TP}$ (true positive), $\mathrm{FP}$ (false positive), and $\mathrm{FN}$ (false negative) denote the proportions of each prediction outcome relative to the total number of analyzed pixels. $E_{\mathrm{base}}$, $E_{\mathrm{forecast}}$, and $E_{\mathrm{perfect}}$ represent the expected expenses when relying on the baseline average inundation map, our ensemble forecast, and an idealized perfect forecast, respectively. The final $\mathrm{REV}$ is calculated using the following equation: 
\begin{equation} 
\mathrm{REV} = \frac{E_{\mathrm{base}} - E_{\mathrm{forecast}}}{E_{\mathrm{base}} - E_{\mathrm{perfect}}}.
\end{equation} 

Furthermore, to evaluate the relative utility of our ensemble forecasts, we conducted a comparative analysis against a deterministic, hypothetical hazard map. For this purpose, we utilized the actual inundation record of the 2011 Tohoku-oki tsunami \cite{gsi2011tsunami}---which is also referenced in the development of the current official hazard map of Sendai city---and computed its corresponding expected expense, $E_{\mathrm{hazardmap}}$, to assess its REV.


\begin{figure*}
    \centering
    \includegraphics[page=1, width=\linewidth]{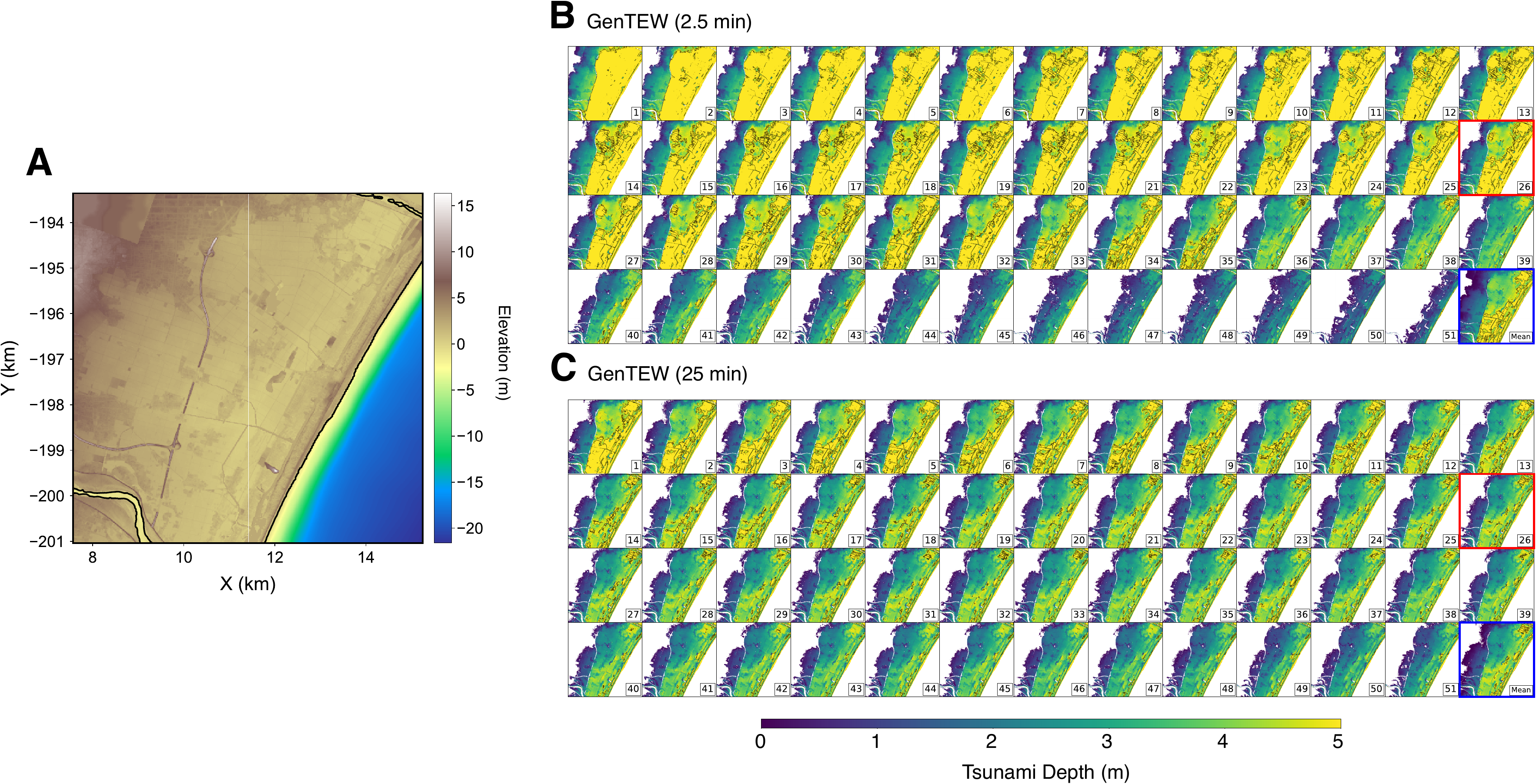}
 
\caption{\textbf{Topography of the target area for inundation prediction and the predicted inundation depths of all 51 ensemble members generated by \textsf{GenTEW}.} \textbf{A}, Topography of the coastal area of Sendai city, the target region for inundation prediction in this study. The contour lines indicate an elevation of 0~m. \textbf{B} and \textbf{C}, Inference results by \textsf{GenTEW} in the scenario based on the estimated source model by Satake et al. \cite{satake2013time} (Fig.~\ref{fig:ensemble_tradeoff}).
In both panels, the prediction maps of all 51 sampled members (Members 1--51) are ordered by total inundation volume, with the median member and the ensemble mean highlighted by red and blue frames, respectively.
\textbf{B}, Predicted inundation depth maps at 2.5 minutes post-event. This illustration explicitly demonstrates the high uncertainty and diverse prediction patterns among members at the initial stage when observation information is critically lacking. \textbf{C}, Predicted inundation depth maps at 25 minutes post-event. It can be visually confirmed that as the observation data accumulate, the independent member predictions converge toward the ground truth distribution (Fig.~\ref{fig:ensemble_tradeoff}C), demonstrating a reduction in prediction uncertainty.}

    \label{fig:ed_ensemble_all}
\end{figure*}

\begin{figure*}
    \centering
    \includegraphics[page=5, width=0.8\linewidth]{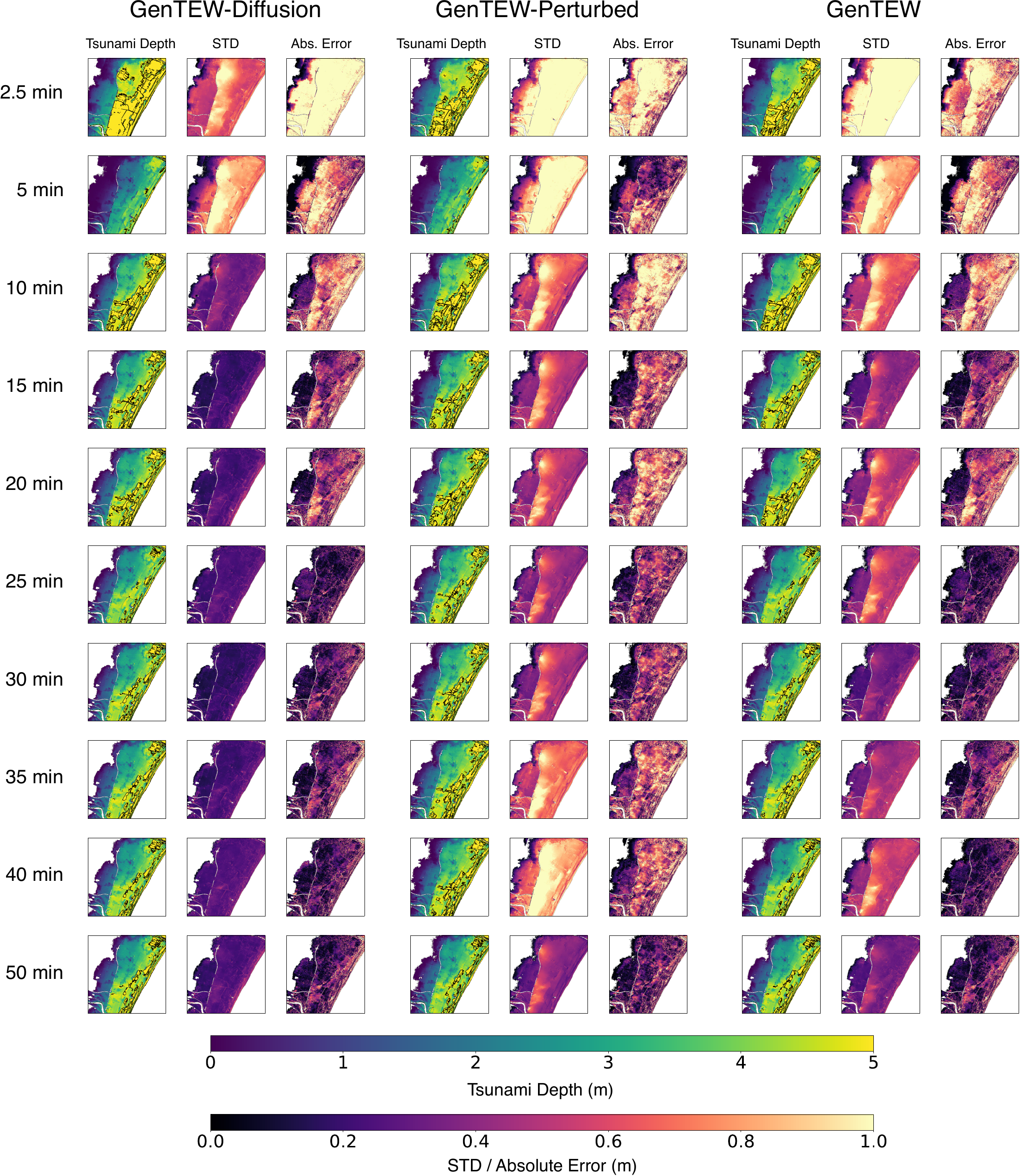}

\caption{\textbf{Full time evolution of the predicted mean inundation depth, standard deviation, and error for each model assuming a high-density observation network (S-net and NOWPHAS) for the 2011 Tohoku-oki tsunami.} Detailed inference results for the scenario shown in Fig.~\ref{fig:ensemble_tradeoff}. For the three models (\textsf{GenTEW-Diffusion}, \textsf{GenTEW-Perturbed}, and \textsf{GenTEW}), the panels show the spatial distributions of the predicted mean inundation depth (Tsunami Depth), ensemble standard deviation (STD), and absolute error (Abs. Error) relative to the target physical simulation result (ground truth) at each time step from 2.5 to 50.0 minutes post-event. As observation data accumulate over time, both the absolute error and the prediction spread (STD) of \textsf{GenTEW} visually decrease, demonstrating prediction convergence. In contrast, \textsf{GenTEW-Diffusion} exhibits small STD values, implying its overconfidence, while \textsf{GenTEW-Perturbed} yields larger overall errors than the other methods.}

    \label{fig:ed_time_evolution_snet}
\end{figure*}

\begin{figure*}
    \centering
    \includegraphics[page=8, width=1.0\linewidth]{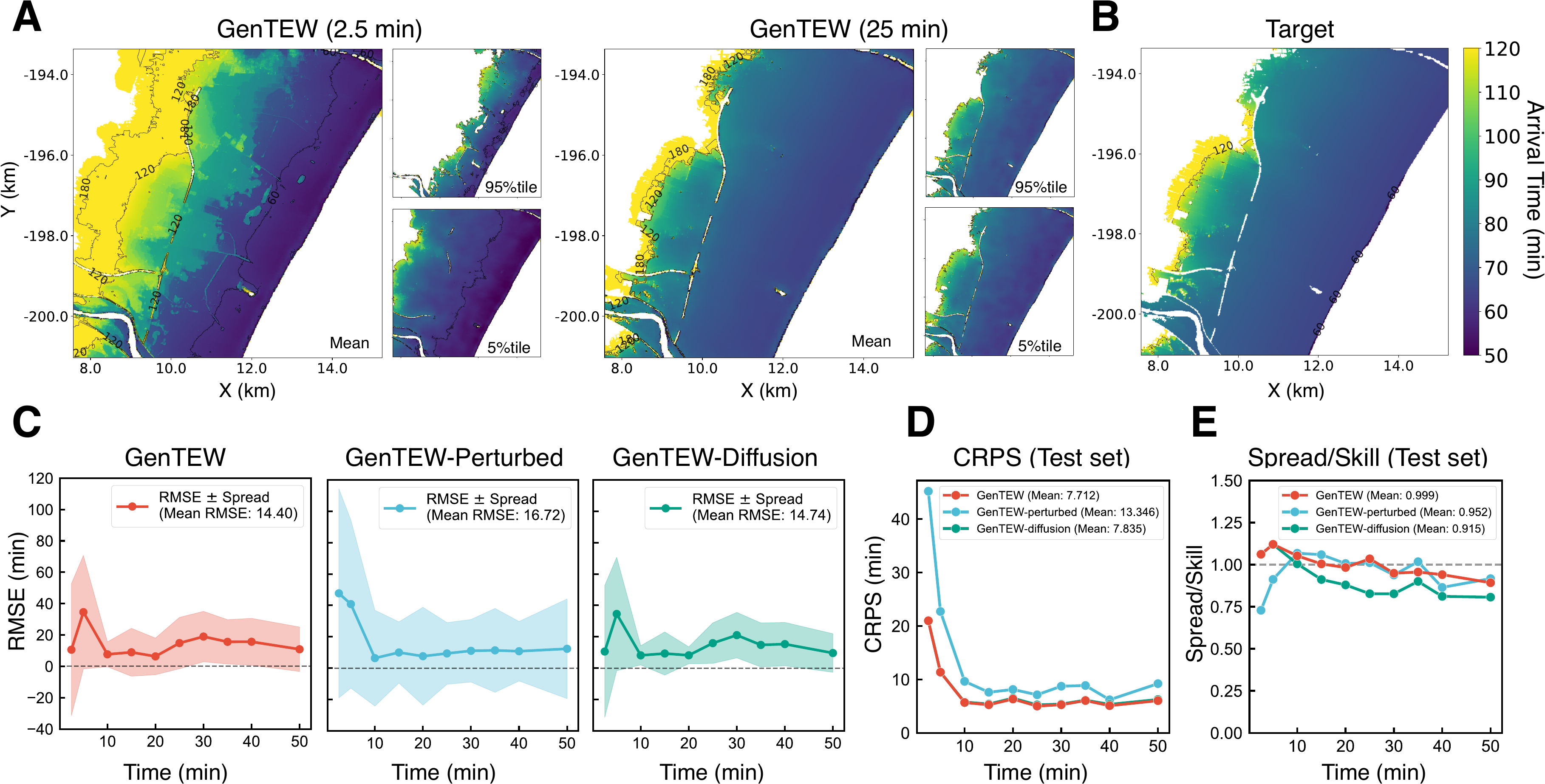}

\caption{\textbf{Quantitative evaluation of \textsf{GenTEW}'s tsunami arrival time predictions using a simulated scenario of the 2011 Tohoku-oki earthquake and test data.}
Evaluation of arrival times for the same 2011 Tohoku-oki tsunami scenario analyzed in Fig.~\ref{fig:ensemble_tradeoff}, using synthesized S-net and NOWPHAS observation data.
\textbf{A}, Predicted mean arrival time maps for Sendai city generated by \textsf{GenTEW} at 2.5 and 25 minutes post-event, alongside inference results corresponding to the 5th and 95th percentiles. \textbf{B}, Ground truth tsunami arrival time map generated by the target physical simulation. At 2.5 minutes post-event, the prediction spread is wide and prediction accuracy of the ensemble mean is limited. By 25 minutes, the predictions visually converge and the mean closely approaches the ground truth. \textbf{C}, Time evolution of the prediction error (RMSE) of the ensemble mean for the three models. The shaded areas indicate the ensemble standard deviation (spread). \textbf{D}, \textbf{E}, Average scores for arrival time prediction across the entire test dataset. \textbf{D}, CRPS. \textbf{E}, Spread-skill Ratio. \textsf{GenTEW} demonstrates that the prediction spread converges within 10 minutes of observation time. Consistent with the inundation depth forecasting results, \textsf{GenTEW-Diffusion} exhibits overconfidence, and \textsf{GenTEW-Perturbed} yields inferior CRPS. \textsf{GenTEW} achieves superior performance in arrival time forecasting compared to the baseline models.}

    \label{fig:ed_arrival_time_stats}
\end{figure*}

\begin{figure*}
    \centering
    \includegraphics[page=9, width=0.8\linewidth]{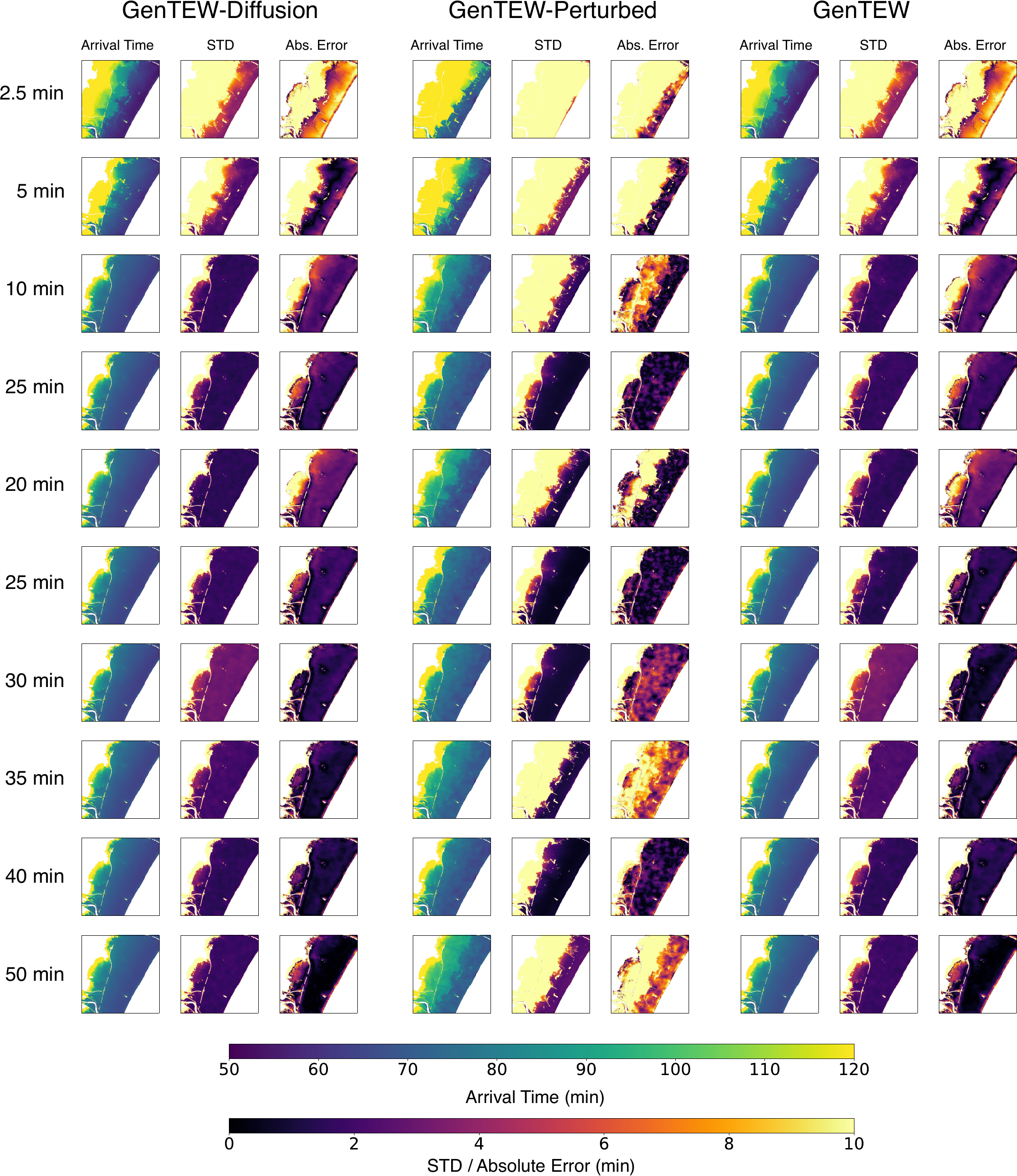}
\caption{\textbf{Full time evolution of the predicted mean arrival time, standard deviation, and error for each model.} Detailed inference results for the arrival time prediction of the 2011 Tohoku-oki tsunami shown in Fig.~\ref{fig:ed_arrival_time_stats}. For the three models (\textsf{GenTEW-Diffusion}, \textsf{GenTEW-Perturbed}, and \textsf{GenTEW}), the panels show the spatial distributions of the predicted mean arrival time (Arrival Time), ensemble standard deviation (STD), and absolute error (Abs. Error) relative to the ground truth target at each time step from 2.5 to 50.0 minutes post-event. \textsf{GenTEW} and \textsf{GenTEW-Diffusion} exhibit prediction convergence to high accuracy at around 10 minutes post-event, while \textsf{GenTEW-Perturbed} shows unstable predictions.}

    \label{fig:ed_time_evolution_arrival}
\end{figure*}

\begin{figure*}
    \centering
    \includegraphics[page=6, width=0.6\linewidth]{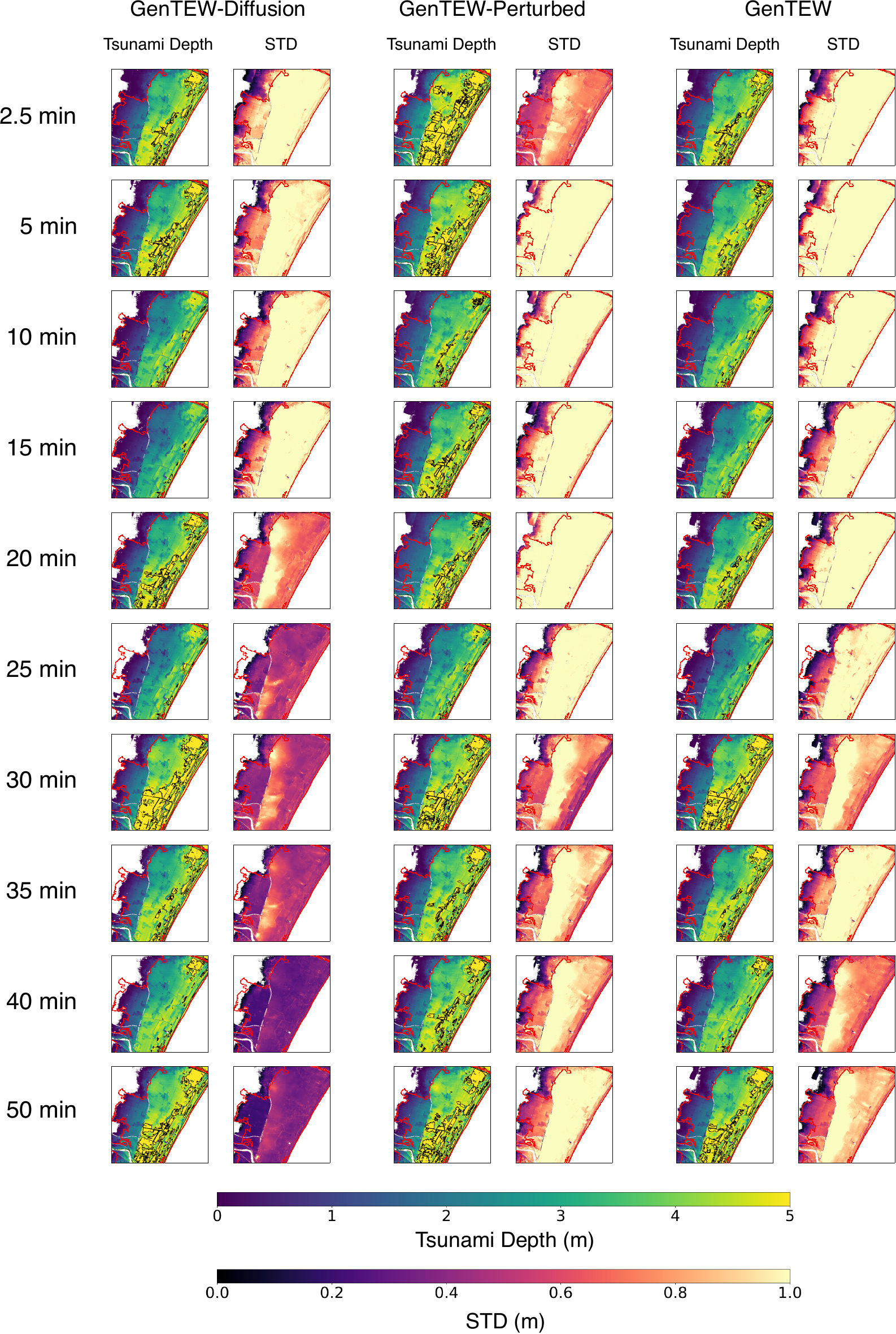}
\caption{\textbf{Full time evolution of the predicted mean inundation depth and standard deviation for each model under a sparse observation network (NOWPHAS only).} Detailed inference results for the same empirical evaluation scenario analyzed in Fig.~\ref{fig:gentew_performance_2011}, using actual observation data (NOWPHAS only) from the 2011 Tohoku-oki earthquake. For the three models (\textsf{GenTEW-Diffusion}, \textsf{GenTEW-Perturbed}, and \textsf{GenTEW}), the panels show the spatial distributions of the predicted mean inundation depth (Tsunami Depth) and the ensemble standard deviation (STD) at each time step from 2.5 to 50.0 minutes post-event. The actual surveyed inundation extent is indicated by the red polygons \cite{gsi2011tsunami}. The spread (STD) of \textsf{GenTEW} and \textsf{GenTEW-Perturbed} decreases at around 30.0 minutes, demonstrating their prediction convergence.}

    \label{fig:ed_time_evolution_nowphas}
\end{figure*}

\begin{table*}
\centering
\caption{\textbf{Optimization results for the multiplicative noise intensity upper limit ($\alpha$) according to the observation network configurations and prediction targets.}
Optimal noise levels ($\alpha$) and statistical evaluation scores (RMSE, spread-skill ratio) evaluated on the validation dataset for three scenarios: \textbf{A}, Inundation depth prediction using the high-density observation networks (S-net and NOWPHAS) analyzed in Fig.~\ref{fig:ensemble_tradeoff}. \textbf{B}, Inundation depth prediction under the sparse observation conditions of the 2011 Tohoku-oki earthquake (NOWPHAS only) analyzed in Fig.~\ref{fig:gentew_performance_2011}. \textbf{C}, Arrival time prediction for the same high-density network scenario as in \textbf{A}. The bold text accompanied by an asterisk ($^\ast$) indicates underdispersed (overconfident) cases where the spread-skill ratio failed to exceed 1.0 even at the maximum search limit ($\alpha=5.0$). The bold text accompanied by a dagger ($^\dagger$) indicates cases where the spread-skill ratio reached 1.0 or greater without waveform perturbation ($\alpha=0.0$), rendering additional noise application unnecessary. Note that for the baseline method \textsf{GenTEW-Diffusion}, which does not apply noise perturbations to the input waveforms, $\alpha$ remains 0.0 across all time periods.}
\label{tab:optimization_results}
\footnotesize
\begin{tabular}{l @{\hspace{1.5em}} ccc @{\hspace{2.5em}} ccc @{\hspace{2.5em}} ccc}
\hline
& \multicolumn{3}{c@{\hspace{2.5em}}}{\textbf{GenTEW-Perturbed}} 
 & \multicolumn{3}{c@{\hspace{2.5em}}}{\textbf{GenTEW}} 
 & \multicolumn{3}{c}{\textbf{GenTEW-Diffusion}} \\
Time & Noise $\alpha$ & RMSE & Spread/ & Noise $\alpha$ & RMSE & Spread/ & Noise $\alpha$ & RMSE & Spread/ \\
(min) & & & Skill & & & Skill & & & Skill \\
\hline
\multicolumn{10}{l}{\textbf{(A) Inundation Depth Prediction with S-net and NOWPHAS (RMSE in meters)}} \\
\hline
2.5 & \textbf{5.00}$^\ast$ & 1.950 & 0.808 & \textbf{5.00}$^\ast$ & 1.863 & 0.907 & 0.00 & 2.127 & 0.543 \\
5.0 & 0.31 & 0.835 & 1.001 & \textbf{0.00}$^\dagger$ & 0.697 & 1.145 & 0.00 & 0.697 & 1.145 \\
10.0 & 0.49 & 0.691 & 1.002 & 0.30 & 0.526 & 1.027 & 0.00 & 0.538 & 0.660 \\
15.0 & 0.52 & 0.567 & 1.045 & 0.46 & 0.533 & 1.046 & 0.00 & 0.567 & 0.486 \\
20.0 & 0.65 & 0.586 & 1.061 & 0.51 & 0.544 & 1.057 & 0.00 & 0.557 & 0.545 \\
25.0 & 0.49 & 0.578 & 1.062 & 0.37 & 0.514 & 1.015 & 0.00 & 0.550 & 0.485 \\
30.0 & 0.46 & 0.516 & 1.097 & 0.32 & 0.454 & 1.000 & 0.00 & 0.479 & 0.543 \\
35.0 & 0.66 & 0.672 & 1.077 & 0.34 & 0.501 & 1.033 & 0.00 & 0.505 & 0.605 \\
40.0 & 0.68 & 0.757 & 1.012 & 0.36 & 0.484 & 1.086 & 0.00 & 0.511 & 0.592 \\
50.0 & 0.55 & 0.558 & 1.046 & 0.38 & 0.474 & 1.007 & 0.00 & 0.486 & 0.587 \\

\hline
\multicolumn{10}{l}{\textbf{(B) Inundation Depth Prediction with NOWPHAS (RMSE in meters)}} \\
\hline
2.5 & 2.00 & 2.320 & 0.473 & 2.50 & 1.945 & 0.809 & 0.00 & 1.845 & 0.741 \\
5.0 & \textbf{5.00}$^\ast$ & 2.280 & 0.831 & 3.01 & 1.959 & 1.013 & 0.00 & 2.119 & 0.492 \\
10.0 & \textbf{5.00}$^\ast$ & 1.621 & 0.972 & 1.61 & 1.393 & 1.038 & 0.00 & 1.323 & 0.715 \\
15.0 & 1.32 & 1.512 & 1.001 & 0.60 & 1.256 & 1.004 & 0.00 & 1.400 & 0.655 \\
20.0 & 4.00 & 1.927 & 1.006 & 0.67 & 1.334 & 1.038 & 0.00 & 1.392 & 0.585 \\
25.0 & 0.45 & 1.077 & 1.001 & 0.19 & 0.867 & 1.015 & 0.00 & 0.912 & 0.747 \\
30.0 & 0.96 & 0.859 & 1.039 & 0.48 & 0.787 & 1.046 & 0.00 & 0.793 & 0.794 \\
35.0 & 0.20 & 0.600 & 1.083 & 0.16 & 0.618 & 1.066 & 0.00 & 0.652 & 0.662 \\
40.0 & 0.26 & 0.645 & 1.039 & 0.17 & 0.583 & 1.006 & 0.00 & 0.587 & 0.674 \\
50.0 & 0.40 & 0.721 & 1.015 & 0.25 & 0.608 & 1.001 & 0.00 & 0.643 & 0.629 \\

\hline
\multicolumn{10}{l}{\textbf{(C) Tsunami Arrival Time Prediction with S-net and NOWPHAS (RMSE in min)}} \\
\hline
2.5 & \textbf{5.00}$^\ast$ & 70.261 & 0.839 & \textbf{0.00}$^\dagger$ & 47.251 & 1.106 & 0.00 & 47.251 & 1.106 \\
5.0 & 0.75 & 48.296 & 1.013 & \textbf{0.00}$^\dagger$ & 34.984 & 1.234 & 0.00 & 34.984 & 1.234 \\
10.0 & 0.50 & 35.142 & 1.002 & 0.10 & 26.569 & 1.002 & 0.00 & 26.708 & 0.934 \\
15.0 & 0.56 & 28.897 & 1.038 & 0.17 & 25.552 & 1.008 & 0.00 & 26.180 & 0.904 \\
20.0 & 0.59 & 29.868 & 1.025 & 0.21 & 27.336 & 1.005 & 0.00 & 27.838 & 0.916 \\
25.0 & 0.44 & 29.285 & 1.007 & 0.22 & 24.979 & 1.013 & 0.00 & 26.215 & 0.807 \\
30.0 & 0.46 & 28.892 & 1.004 & 0.16 & 24.706 & 1.010 & 0.00 & 25.553 & 0.888 \\
35.0 & 0.56 & 31.385 & 1.021 & 0.10 & 26.394 & 1.001 & 0.00 & 26.566 & 0.953 \\
40.0 & 0.36 & 22.215 & 1.007 & 0.17 & 22.365 & 1.015 & 0.00 & 22.796 & 0.903 \\
50.0 & 0.60 & 30.666 & 1.027 & 0.15 & 25.398 & 1.025 & 0.00 & 25.840 & 0.924 \\

\hline
\end{tabular}
\end{table*}

\clearpage
\end{document}